%% file: acl_latex.tex
\pdfoutput=1

\documentclass[11pt]{article}

\usepackage[preprint]{acl}

\usepackage{times}
\usepackage{latexsym}
\usepackage{enumitem}
\usepackage{placeins}
\usepackage[T1]{fontenc}

\usepackage[utf8]{inputenc}

\usepackage{microtype}

\usepackage{inconsolata}

\usepackage{graphicx}
\usepackage{amssymb}
\usepackage{amsmath}
\usepackage{booktabs}
\usepackage{soul}
\usepackage{pifont}
\newcommand*{\revision}{} 

 \newcommand*{\remove}[1]{\leavevmode\ignorespaces} 
\newcommand{\cmark}{\ding{51}}%
\newcommand{\xmark}{\ding{55}}%

\makeatletter
\newcommand\footnoteref[1]{\protected@xdef\@thefnmark{\ref{#1}}\@footnotemark}
\makeatother

\title{Discovering Properties of Inflectional Morphology\\in Neural Emergent Communication}

\author{Miles Gilberti \hspace{40pt} Shane Storks \hspace{40pt} Huteng Dai \\
       University of Michigan \\
        \texttt{\{milgil,sstorks,huteng\}@umich.edu} \\  }

\begin{document}
\maketitle
\begin{abstract}
Emergent communication (EmCom) with deep neural network-based agents promises to yield insights into the nature of human language, but remains focused primarily on a few subfield-specific goals and metrics that prioritize communication schemes which represent attributes with unique characters one-to-one and compose them syntactically. We thus reinterpret a common EmCom setting, the attribute-value reconstruction game, by imposing a small-vocabulary constraint to simulate double articulation, and formulating a novel setting analogous to naturalistic inflectional morphology (enabling meaningful comparison to natural language communication schemes). We develop new metrics and explore variations of this game motivated by real properties of inflectional morphology: concatenativity and \revision{fusion}. Through our experiments, we discover that simulated phonological constraints encourage concatenative morphology, and emergent languages replicate the tendency of natural languages to fuse grammatical attributes.
\end{abstract}

\begin{figure*}[t]
    \centering
    \includegraphics[width=0.95\linewidth,trim={0 0 0 2.9cm},clip]{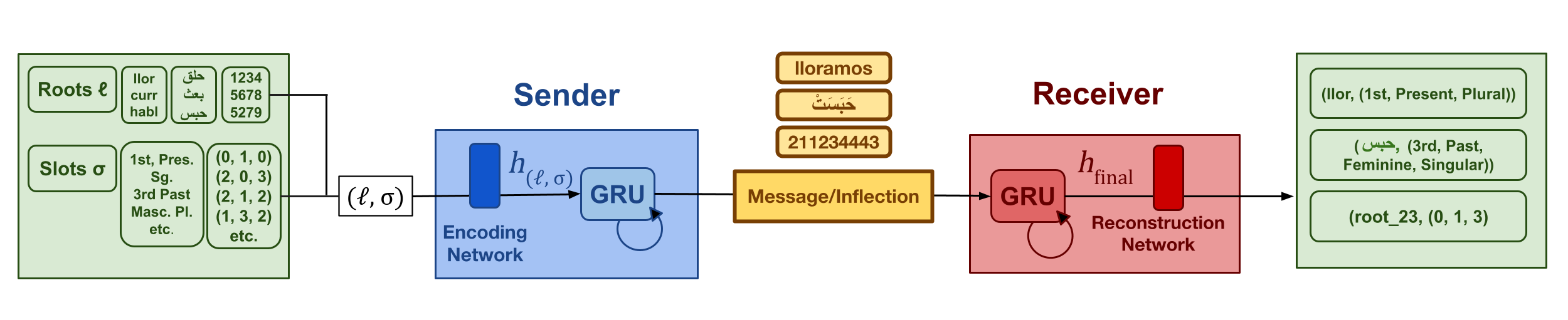}                               
    \vspace{-12pt}
    \caption{We reframe the typical attribute-value (Attr-Val) reconstruction game with an analogy to inflectional morphology, where roots $\ell$ and slots $\sigma$ (e.g., tense and person) comprise attributes which must be communicated through messages. Emergent communication schemes are then comparable to natural language inflections.} 
    \vspace{-10pt}
    \label{fig:overview}
\end{figure*}
\input{sections/1-introduction}
\input{sections/2-related_work}
\input{sections/3-problem_statement}
\input{sections/4-methods}
\input{sections/5-results}

\input{sections/6-conclusion}
\input{sections/7-etc}

\bibliography{custom}

\input{sections/8-apx}

\end{document}

%% file: sections/1-introduction.tex
\section{Introduction}

Emergent communication (EmCom) studies communication protocols developed between two or more \revision{deep neural} agents engaged in a task requiring successful communication. 
Analyses of learned protocols in EmCom typically search for characteristics deemed crucial in natural language, prominently compositionality \cite{chaabouni-etal-2020-compositionality}.
Such experiments promise to shed light on the pressures underlying human-like language evolution. 

However, as interpreting emergent communication schemes and measuring compositionality are challenging, inhibitive assumptions are often made.
First, most EmCom works afford agents an over-complete vocabulary of characters, making it possible for agents to describe each possible attribute value with a unique character one-to-one, thus leading to treatment of generated characters exclusively as words composed only syntactically.
In turn, common evaluation metrics developed for this large-vocabulary setting are overly simplified, 
favoring protocols that represent attribute values with consistent characters concatenated in a consistent ordering \cite{chaabouni-etal-2020-compositionality,resnick,ueda2023on}.
This disregards the possibility of many morphological phenomena attested in natural languages, such as nonconcatenative morphology (e.g., root-pattern morphology common in Semitic languages), and fusion of grammatical attributes in morphemes (e.g., the Spanish verb suffix \textit{-amos} simultaneously communicates first person, plurality, and present tense).

In this work, we reinterpret EmCom as \textit{emergent morphology}, addressing these shortcomings with two key design choices. First, we target humanlike \textit{double articulation \cite{hockett} by enforcing a small character inventory, necessitating that individually meaningless characters form combinatorial morpheme-like symbols to encode attribute values and combinations thereof.}
Second, as shown in Figure~\ref{fig:overview}, we introduce a novel configuration of attribute-value reconstruction inspired by inflectional morphology, where a high-cardinality attribute (analogous to roots) is paired with comparatively lower-cardinality attributes (analogous to grammatical information, e.g., tense and person), encouraging richer morphological phenomena.
We review EmCom evaluation metrics for various properties of communication schemes, proposing novel metrics for the underexplored properties of concatenativity and \revision{degree of fusion}. To validate and establish expected ranges for these metrics, we then analyze a representative group of artificial double articulatory communication schemes and inflection-based natural language communication schemes.

Lastly, we leverage this reimagined problem to explore novel, linguistically compelling research questions. We implement a phonology-inspired constraint in communication agents, using concatenativity metrics to reveal ease of articulation as a pressure toward concatenative languages. We then compare the topographic similarity of emergent and natural languages, finding significant room for improvement in emergent language to reflect natural degrees of compositionality. Finally, we analyze the presence of fusion in emergent and natural languages for inflection, finding that emergent languages most often fuse grammatical attributes rather than roots, mirroring natural language. \remove{Altogether, }\revision{This work opens exciting new directions for} future research into the evolution of morphology.

%% file: sections/2-related_work.tex
\section{Related Work}
\revision{Some other EmCom works address double articulation explicitly via selection of small inventory sizes \cite{ueda-washio-2021-relationship,ueda2023on,DBLP:conf/iclr/UedaT24}, but the analysis focuses on words rather than morphemes, and only considers perfectly concatenative methods of combining them.
Work utilizing bitstring messages in EmCom achieves this \cite{resnick,gupta-etal-2020-compositionality}, but not in a linguistically realistic way.}

\revision{Unlike our work, many works use large-vocabulary EmCom to investigate various phenomena relevant to natural language morphology.
For example, \citet{NEURIPS2019_31ca0ca7} use a length pressure to rederive Zipf's Law of Abbreviation\remove{, a statistical property of natural language}.
\citet{lian-etal-2023-communication,lian-etal-2024-nellcom} explore the 
case-marking versus word-order tradeoff in pair and group communication.
\citet{conklin2023compositionality} discover naturalistic variability in emergent languages through proposed metrics for synonymy, homonymy, word-order freedom, and entanglement, complicating typical measures of compositionality. 
\citet{galke2022emergentcommunicationunderstandinghuman,galke_raviv} explore the importance of learnability, generalization, production effort, and group size pressures, 
among other psycho-sociolinguistic factors. 
\citet{chaabouni2022emergent} further explore the relationships between these aspects in a naturalistic visual task setting, while \citet{kouwenhoven-etal-2024-curious} investigate the effect of agents' representational alignment on their performance.}

\revision{Lastly, there is vast literature studying morphology in human languages. Notably, we adopt the term ``emergent morphology'' introduced by \citet{archangeli2016emergent} within their \textit{Emergent Grammar} framework, which hypothesizes how morphophonological structures arise from constraint interactions in natural languages. Especially related prior works attempt to model and analyze the emergence of various aspects of morphological systems \cite{nowak,zuidema,elsner2020stop,dekker}. Unlike these works, we emphasize computational modeling of \textit{artificial communication systems}, focusing on how minimal meaningless discrete units combine systematically into meaningful morphological structures like inflection. Automated morphological segmentation, especially for nonconcatenative morphology \cite{fullwood2018biases}, is also relevant. While we only explored concatenative approaches, we expect such techniques to play a crucial role in future work toward understanding morphology in EmCom.}

%% file: sections/3-problem_statement.tex
\section{Toward Emergent Morphology}


Prior \revision{EmCom work} has largely focused on the Lewis Signaling Game \cite{kottur-etal-2017-natural,chaabouni-etal-2020-compositionality,ueda2023on, lewis-signalling-game}, in which a sender agent observes some state and \revision{generates} a message to a receiver agent, which must take an action based on \revision{it}. Success in the game relies on \revision{accurately communicating} the relevant features of the environment. 
\revision{Morphology has been relatively understudied in EmCom due to a focus on individual characters representing atomic units of meaning \cite{boldt2024a}.}

\subsection{Attribute-Value Reconstruction Game}

\revision{The Attrbute-Value Reconstruction Game (henceforth \textbf{Attr-Val})} has been extensively studied by EmCom researchers \cite{kottur-etal-2017-natural,chaabouni-etal-2020-compositionality}. This class of games models the state as an $n$-tuple of values where each value belongs to some attribute $A_i \in \mathcal{A}$, defined as a set of possible values for the attribute. That is, the set of states is $S = \{s|s =(v_1 \in A_1, v_2 \in A_2, \dots, v_n \in A_n)\}$. 

Formally, \remove{an instance of }Attr-Val can be defined as such: we have a state $s \in S$, a sender $f_\theta: S\to M$, and a listener $g_\theta: M\to S$. $M$ is the message space, which consists of messages of up to length $m$ characters, all of which must belong to a fixed character vocabulary $C$. This instance is considered successful if $g_\theta(f_\theta(s)) = s$. 
Following prior works, we use EGG \cite{egg} to implement sender and receiver agents as single layer gated recurrent unit (GRU; \citealp{gru}) models.\footnote{\label{fn:agents}\revision{More details about agents provided in Appendix~\ref{apx:agent impl}.}}

\subsection{\revision{Attr-Val as Inflectional Morphology}}

The attribute value sets $A \in \mathcal{A}$ tend to have the same cardinality in prior work \cite{kottur-etal-2017-natural,chaabouni-etal-2020-compositionality,ueda2023on}\revision{, often motivating analogies to communicating properties of objects with few possible values, e.g.,} a red square or blue triangle. 
\revision{We claim that Attr-Val games also represent a rich analogy for inflectional morphology. }The meaning of an inflected root in natural language consists of the root as well as additional attributes like tense and person for verbs, or number and gender for nouns. Like Attr-Val, these attributes take on a finite set of 
values.
While the number of possible roots is unbounded in natural language, Attr-Val can simulate this through one attribute set (representing roots) being much larger than others. This paradigm provides a key advantage: natural languages solve an analogous and similarly challenging problem to this, 
enabling direct comparison with learned communication schemes. 

\revision{We consider 4 configurations for $\mathcal{A}$: a \textbf{default} setting resembling prior work with 3 attributes of 16 values each $(16 \times 16 \times 16)$, and 3 novel \textbf{inflection} settings with nonuniform attribute value sets $(42 \times 2 \times 3)$, $(42 \times 2 \times 2 \times 2)$, and $(42 \times 6)$. In inflection settings, the largest attribute set symbolizes roots, while others symbolize grammatical properties like tense and person.}
\remove{Both }Settings have a comparable total number of attribute values \revision{(47-48)}.\footnote{\revision{We also ran experiments controlling for total number of combinations, but models failed to converge (see Appendix~\ref{apx:inflection note}).}} To necessitate double articulation, i.e., \revision{composing meaningless characters into meaningful symbols} to represent attributes, we enforce a consistent character vocabulary size $|C|=8$ and message length $m=9$, smaller than the size of most attribute sets. 
\revision{Importantly, one attribute in the inflection settings has many more possible values than $|C|$ while others have fewer, possibly incentivizing more morphologically diverse communication schemes.}





%% file: sections/4-methods.tex
\section{Interpreting Communication Schemes}

To investigate the emergence of double articulation in EmCom, we next introduce methods to evaluate communication schemes by extracting meaningful symbols from messages, and measuring various properties of them.
\remove{Toward establishing a typology of possible compositional communication schemes, }We then establish reference points for emergent communication schemes based on both idealized artificial languages and natural languages.

\subsection{Evaluating Communication Schemes}

The need for stronger evaluations in EmCom is well-established \cite{boldt2024a,chaabouni-etal-2020-compositionality}, and especially prominent in our problem setting, which aims to consider a greater number of plausible communication schemes than prior work. As such, we next introduce and appraise several algorithms for segmenting meaningful symbols from sequences of characters, as well as evaluation metrics for sequences of symbols in communication schemes.

\subsubsection{Symbol Segmentation Algorithms}
To extract meaningful, morpheme-like symbols from messages (essential for double articulation), we consider two approaches: one from prior work based on Harris' articulation scheme \cite{ueda2023on}, and a previously unexplored approach based on byte-pair encoding \cite{bpe}. We define the resulting segmented symbol vocabulary $V$ as the set union of symbols extracted from each message.

\paragraph{Harris' Articulation Scheme}
Harris' articulation scheme (HAS) has been proposed as a desirable, naturalistic trait for emergent languages to exhibit \cite{ueda2023on,DBLP:conf/iclr/UedaT24}. HAS captures the tendency of word boundaries to be identifiable purely by character-level entropy. Specifically, the scheme claims that while entropy generally decreases across an utterance, it will increase at morpheme boundaries. This implies an algorithm for segmenting morphemes by drawing boundaries at indices $i$ where $\mathcal{H}_i -\mathcal{H}_{i+1} > \tau$, where $\tau$ is a threshold that determines how sensitive the algorithm is. In all of our analyses, $\tau$ is set to 0 to maximize sensitivity. We use the implementation provided by \citet{ueda2023on}.


\paragraph{Byte-Pair Encoding}
HAS is a powerful tool for identifying symbols in communication schemes, but it is difficult to control the degree of segmentation by it.
As such, we propose an alternative approach based on the Byte-Pair Encoding (BPE) algorithm \cite{bpe}, which has been widely used in tokenizers for NLP systems \cite{sennrich-etal-2016-neural}.
Importantly for our purposes, the properties of the subwords generated by BPE have been shown to enable characterization of languages by established typological categories, such as analytic and synthetic \cite{10.1162/coli_a_00489}.
BPE is configurable with a maximum vocabulary size $|V|$. Early compressions tend to be most informative \cite{gutierrez-vasques-etal-2021-characters,10.1162/coli_a_00489}, so we apply BPE with a merging cutoff of 96, approximately twice the expected number of symbols (i.e., the total number of unique attribute values), and additionally apply BPE with maximum compression.







\subsubsection{Evaluation Metrics}\label{sec:metrics}
Given a segmented symbol vocabulary $V$, we next \revision{discuss} common metrics for its compositionality from prior work\revision{. We then introduce} methods to evaluate concatenativity and \revision{degree of fusion}, more specific forms of compositionality that typical compositionality metrics do not directly measure\revision{, as well as learnability, which has previously been connected to compositionality \cite{KIRBY2014108,NEURIPS2019_b0cf188d,chaabouni-etal-2020-compositionality}.}

\paragraph{Compositionality}
Compositionality is the core objective for many EmCom works, motivated by interpretability and its prominence in natural languages. 
Topographic similarity (\textbf{TopSim}) is a common measure for this \cite{topsim,lazaridou2018emergence}, formally defined as the correlation of the distance (typically Levenshtein distance; \citealp{Levenshtein1965BinaryCC}) between pairs of messages and the distance between their meanings. 
This provides a global measure of how similar the symbols in messages for similar configurations.

However, TopSim places minimal restrictions on how similarity of messages is calculated. As such, \citet{chaabouni-etal-2020-compositionality} propose bag-of-symbols disentanglement (\textbf{BoSDis}), an alternative metric accounting for permutation-invariant communication schemes\footnote{They also propose positional disentanglement, which we omit but provide more details about in Appendix~\ref{apx:posdis}.} \revision{(e.g., ``A and B'' versus ``B and A'')}:

\vspace{-6pt}

\begin{equation}
    \frac{1}{|V|}\sum_{v\in V}\frac{\mathcal{I}(n_v; a_v) - \mathcal{I}(n_v; b_v)}{\mathcal{H}(n_v)}
\end{equation}

BoSDis calculates the \revision{mean entropy-normalized} difference in mutual information $\mathcal{I}$ between the count $n_v$ of a vocabulary symbol $v \in V$ and attributes $a_v$ and $b_v$, which \revision{have the 2 highest $\mathcal{I}$ values with $v$. }
\remove{This is normalized by entropy $\mathcal{H}$ for $n_v$. }
This \remove{metric }\revision{measures whether} a \revision{character's} presence in a message uniquely communicates a specific \revision{attribute value}. This should generally not happen in our small-vocabulary setting, which necessitates reuse of characters in multi-character symbols to induce double articulation,\footnote{\revision{One-character symbols are an exception, but an accurate communication scheme has very limited capacity for these.}} thus BoSDis should \remove{generally }be low.
However, if \revision{an algorithm effectively segments symbols} which uniquely refer to \revision{attributes}, the BoSDis with respect to those symbols should be higher. 
We thus \remove{use this as a }gauge \remove{of }whether HAS and BPE extract meaningful symbols through a ratio \textbf{BoSDis$^V_C$} of \remove{the }BoSDis for a segmented symbol vocabulary $V$ \revision{and} unsegmented \revision{characters} $C$. If BoSDis$^V_C > 1$, \revision{then} $V$ creates a more disentangled representation than individual characters $C$, \revision{thus $V$ contains} meaningful multi-character symbols.

\paragraph{Concatenativity}

TopSim and BoSDis fail to account for non-trivial \revision{and variable} forms of compositionality, including some present in natural language \revision{\cite{korbak2020measuring,conklin2023compositionality}}. For example, TopSim based on Levenshtein distance of characters reflects an assumption that characters should not be modified or reordered under composition. Contrarily, we aim to consider \revision{any scheme where} messages are formed via a systematic composition of symbols (i.e., as opposed to a memorized ``hash function'') as compositional.

One particular form of compositionality we aim to tease apart from others is \textit{concatenativity}, i.e., the degree to which composable symbols consist of uninterrupted streams of characters (analogous to phonemes) that are concatenated under composition.\footnote{\revision{This is closely related to \textit{integrity} in optimality theory, which prohibits output forms from containing multiple correspondents to an input form \cite{faithfulness_identity}.}}
Using English verb conjugation as an example, ``played'' is a concatenative message consisting of ``play'' and ``ed.'' By contrast, ``sang'' is a nonconcatenative message since the characters ``s\_ng'' correspond to the root ``sing,'' but are interrupted by ``a,'' which communicates tense.

As HAS and BPE require symbols to be composed of consecutive characters, we estimate concatenativity as the average number of symbols in segmented messages of a communication scheme.
We refer to these length metrics as \textbf{HASLen} and \textbf{BPELen$_{|V|}$}.\footnote{For \revision{HASLen}, we use the number of boundaries\remove{ drawn}. \remove{Note that concatenativity implies compressibility via algorithms like BPE, but not necessarily vice versa. }Since HAS relies on entropy, \revision{we note that HASLen may be} artificially low in communication schemes that include some random or otherwise non-meaningful characters.}  A smaller number of symbols indicates greater concatenativity, as more consecutive characters belong in self-contained units of meaning.
Of course, not all compositional communication schemes are concatenative.
\revision{In Section~\ref{sec:artificial lang},} we will apply concatenativity metrics to contrived concatenative and nonconcatenative compositional schemes to analyze how these properties relate.

\paragraph{\revision{Degree of Fusion}}

\revision{\textit{Fusion}} is the expression of multiple attributes of meaning within one morpheme or symbol.
Prior work examines this by taking conditional probabilities of surface forms after ablating feature pairs from \revision{data} \cite{rathi2021information,rathi2022explaining}\revision{, but} this requires \revision{computationally expensive training of models for each feature pair}.
The fusion of two attributes $A_1,A_2$ is equivalent to treating them as one attribute $A_{12} = A_1 \times A_2$, for which symbols may be chosen arbitrarily, and a language may be compositional with respect to a particular fused attribute set.
As such, we define a novel metric of fused TopSim (\textbf{F-TopSim}), calculated by the maximum TopSim over the fusions of all attribute pairs.
A language is fusional if $\text{F-TopSim} \geq \text{TopSim}$, so we summarize \revision{the degree of fusion with} F-TopSim $\delta = \text{F-TopSim} - \text{TopSim}$.

\paragraph{\revision{Learnability}}
\revision{To explore its relationship with concatenativity and fusion, we measure learnability by the average number of training \textbf{epochs} that a listener takes to exceed 99\% reconstruction accuracy.}

\section{Artificial and Natural Reference Points}

Next, we curate reference compositional communication schemes. To evaluate \revision{how metrics capture the above properties and establish expected ranges, }we first propose \remove{several contrived }artificial languages to target them. To contextualize EmCom results with real-world communication schemes, we develop additional languages based on natural language inflections.

\subsection{Artificial Languages}\label{sec:artificial lang}
As shown in Figure~\ref{fig:artificial_lang}, we generate 5 artificial languages \revision{with a range of morphological complexity} using message length $m=8$ and vocabulary size $|C|=9$. 
The first 4 types compose \revision{symbols of constant and mixed length with varied levels of concatenativity}. \remove{Notably, }These languages preserve symbols and the ordering of their characters under composition, favored by TopSim. 
The fusion language violates this by assigning symbols to combinations of attribute values rather than individual ones. 
\remove{These languages illustrate a range of morphological complexity. }
\revision{Tables~\ref{tab:artificial language metrics}-\ref{tab:artificial language metrics 2} report metrics for listeners trained on these languages.} We \revision{list} key observations below.\footnote{\revision{More discussion and artificial languages in Appendix~\ref{apx:more art langs}.}}

First, the degree of concatenation \remove{in artificial languages }has minimal effect on learnability, implying {a lack of inductive bias toward concatenation in recurrent listener networks.}
BoSDis$^V_C$ metrics confirm that {HAS and BPE extract more meaningful morphemes from more concatenative languages}, despite all \remove{artificial }languages being compositional. HASLen and BPELen are lower in more concatenative languages, suggesting these metrics effectively reflect concatenativity.
Mixed concatenative languages have higher variance in BPELen, falling along a spectrum between perfectly concatenative and nonconcatenative languages, and random languages have markedly higher asymptotic BPELen than \revision{others}, further evidence of BPE's effectiveness.
\remove{However, }As expected, HAS is less effective for languages with randomness, evidenced by \revision{low HAS BoSDis$^V_C$ and HASLen in mixed concatenative and variable length languages, with the latter also low in random languages.}\footnote{Supplementary results with HASLen in Appendix~\ref{apx:haslen}.}
BoSDis$^V_C$ metrics are \revision{large} for random languages, suggesting an {over-sensitivity to noise} in the absence of other signal.
Lastly, {F-TopSim $\delta$ effectively \revision{reflects fusion in the fusion languages.}

\begin{figure}
    \centering
    \includegraphics[width=0.8\linewidth,trim={0 2cm 0 1.5cm}, clip]{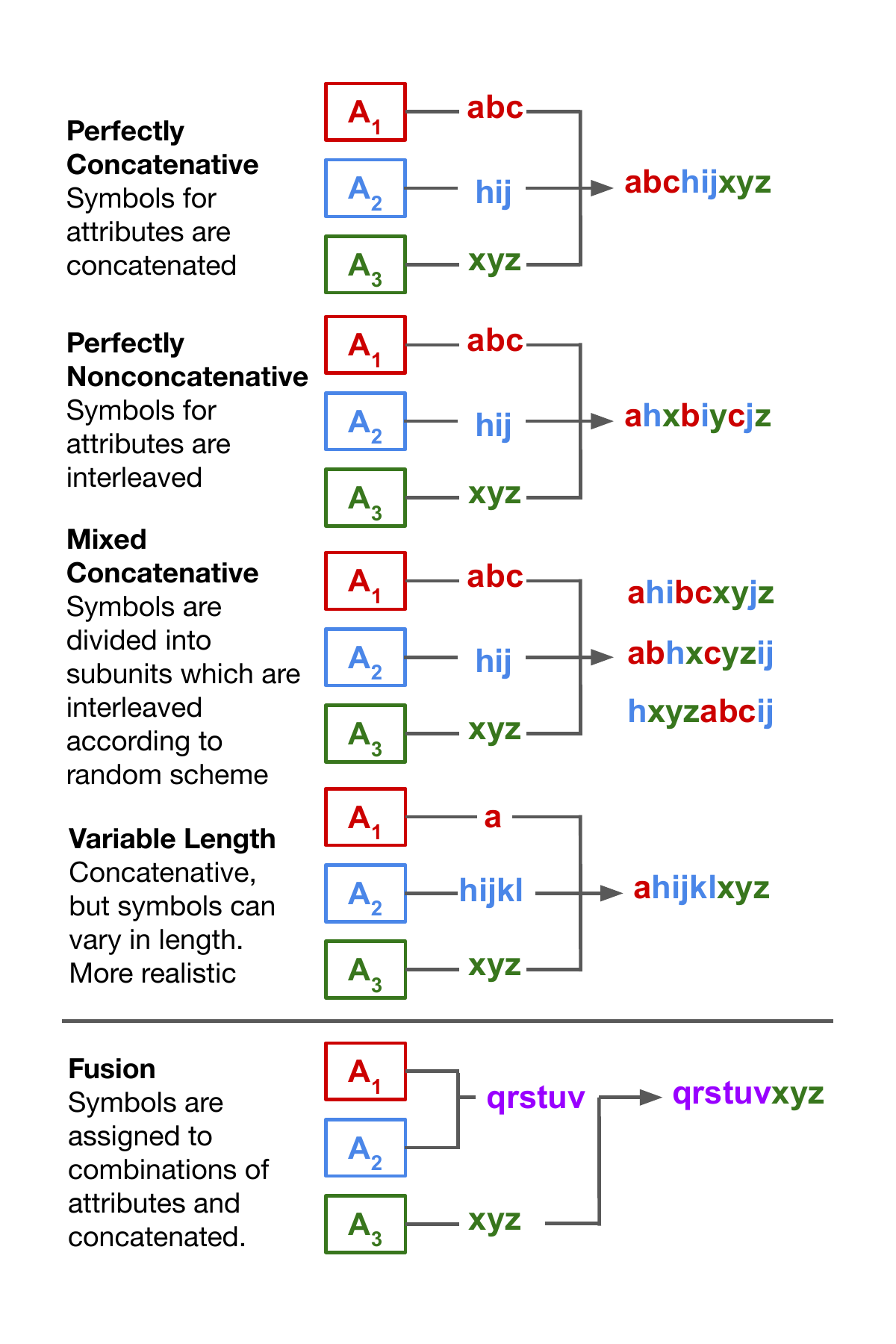}
    \vspace{-5pt}
    \caption{Artificial languages. Each language defines a unique symbol for each possible attribute value, but varies the composition operation for symbols.}
    \vspace{-5pt}
    \label{fig:artificial_lang}
\end{figure}

\begin{table*}[t]
    \centering
    \footnotesize
    \setlength{\tabcolsep}{3.9pt}
    \begin{tabular}{rccccccc} 
        \toprule
         \textbf{Language} &   \textbf{TopSim}&\textbf{BoSDis}&  \textbf{HAS BoSDis$^V_C$}&\textbf{BPE$_{96}$ BoSDis$^V_C$}  &\textbf{BPE$_{\text{Max}}$ BoSDis$^V_C$}  &\textbf{F-TopSim $\delta$}&\textbf{Epochs}\\
         \cmidrule(lr){1-1}\cmidrule(lr){2-6}\cmidrule(lr){7-7}\cmidrule(lr){8-8}
          
 Perf. Conc.&   .628 $\pm .01$&.076$\pm .03$&  4.889$\pm 4.47$&7.836$\pm 3.72$  &.751$\pm$.34&-.236$\pm$.01&4.875$\pm .83$\\ 
 Mixed Conc. &   .621$\pm .00$&.076$\pm .03$&  2.146$\pm .70$&1.829$\pm .84$&.430$\pm$.23&-.236$\pm$.01&3.875$\pm.64$\\ 
 Nonconcat.&   .624$\pm .01$&.076$\pm .03$&  .509$\pm .18$&.825$\pm .43$&.379$\pm$.15&-.237$\pm$.01&4.875 $\pm .99$\\ 
 Var. Length&   .451$\pm .03$&.085$\pm .03$&  1.965$\pm 1.01$&5.110$\pm 1.77$&.796$\pm$.28&-.131$\pm$.02&10.00$\pm3.85$\\
         \cmidrule(lr){1-1}\cmidrule(lr){2-6}\cmidrule(lr){7-7}\cmidrule(lr){8-8}
 Fusion &    .238$\pm .03$&.126$\pm .01$&  9.685$\pm .08$&4.474$\pm .15$&.402$\pm$.03&.175$\pm$.01&29.13$\pm3.09$\\ 
 Random &    .000$\pm .00$&.001$\pm .00$&  3.437$\pm 1.14$&8.263$\pm 2.71$  &19.51$\pm$7.32&.000$\pm$.00&100+\\
 
 \bottomrule
 \end{tabular}
    \normalsize
\vspace{-3pt}

    \caption{Evaluation of compositionality, \revision{degree of fusion}, and learnability \remove{(i.e., average number of training epochs that a listener agent takes to exceed 99\% reconstruction accuracy) }of artificial languages, averaged over 8 random seeds. 1$\sigma$ confidence intervals provided to within 2 decimal points, or whole numbers for \revision{larger} values.}
    \vspace{-10pt}
    \label{tab:artificial language metrics}
\end{table*}

\begin{table}[t]
    \centering
    \footnotesize
    \setlength{\tabcolsep}{4.5pt}
    \begin{tabular}{rccc} 
        \toprule
         \textbf{Language} &\textbf{HASLen}&\textbf{BPELen$_{96}$}&  \textbf{BPELen$_{\text{Max}}$} \\
         \cmidrule(lr){1-1}\cmidrule(lr){2-4}
          
 Perf. Conc.&2.758$\pm .22$&3.185$\pm.06$& 2.001$\pm .00$\\ 
 Mixed Conc. &3.791$\pm .57$&4.489$\pm.20$& 2.178$\pm .15$\\ 
 Nonconcat.&4.558$\pm .32$&4.722$\pm$.07& 2.294$\pm .05$\\ 
 Var. Length&2.463$\pm .30$&3.078$\pm$.18& 1.995$\pm .00$\\
\cmidrule(lr){1-1}\cmidrule(lr){2-4}
Fusion &3.502$\pm .21$&4.341$\pm$.02&  2.004$\pm .01$\\ 
 Random &1.593$\pm .19$&4.997$\pm.01$& 3.105$\pm .01$\\
 
 \bottomrule
 \end{tabular}
    \normalsize
\vspace{-3pt}
    
    \caption{\revision{Artificial language concatenativity metrics} averaged over 8 random seeds\revision{, and} 1$\sigma$ confidence intervals\remove{ provided to within 2 decimal points}.}
    \vspace{-8pt}
    \label{tab:artificial language metrics 2}
\end{table}

\subsection{Natural Languages}

As a comparison to humanlike communication schemes, we also analyze subsets of the verb conjugations of two natural languages, Spanish and Arabic, as Attr-Val languages. We chose these due to the availability of high-quality datasets, the fact that both conjugate for tense and person, and the fact that they represent fairly divergent communication schemes along the concatenativity axis.
Spanish data comes from \remove{the Jehle Spanish Verb Database }\revision{\citet{jehle-verb-database}}, and Arabic data \remove{was sourced }from \citet{nawaz-etal-2025-automated}.
We select a set of verb attributes comparable to the inflection Attr-Val setting: root, tense, and person. We then generate Attr-Val configurations based on combinations of these verb attributes sampled from the data, which serve as cognitively plausible (albeit simplified) representations of actual cognitive objects and events encoded and reconstructed in natural communication. For other attributes beyond tense and person, we choose a constant value (e.g., mood is always indicative). 
We generate 50 such inflectional sublanguages for Spanish and Arabic, each of which are
mappings between lexeme-slot tuples $(\ell,\sigma)$ and surface forms $w$ \cite{wu-etal-2019-morphological}.

\remove{We find that }Up to 76\% of generated Spanish sublanguages can be meaningfully segmented by either HAS or BPE (i.e., BoSDis$^V_C$ > 1), while only up to 8\% of the Arabic sublanguages can be\footnote{Detailed analysis in Appendix~\ref{apx:segment meaningfulness}.}\revision{, demonstrating} how both tools for symbol segmentation are insufficient for nonconcatenative morphology. Other metrics for these sublanguages are discussed in Section~\ref{sec:experiments}.

%% file: sections/5-results.tex
\section{EmCom \revision{Experiments}}\label{sec:experiments}

Our \remove{reimagined} Attr-Val settings targeting double articulation and inflection, general and specific evaluation metrics for communication schemes, and illustrative artificial and natural language reference points enable us to ask previously unexplored research questions around morphology in EmCom. To begin this exploration, we specifically ask:

\begin{enumerate}[noitemsep,nolistsep]
    \item \textit{Do more concatenative languages emerge under a human-inspired evolutionary pressure for ease of articulation?}
    \item \textit{How do topographic similarity in emergent and natural languages for inflection compare?}
    \item \textit{Do emergent languages exhibit \revision{fusion} in inflection similarly to natural languages?}
\end{enumerate}

\revision{We next introduce experimental details, including the above articulation pressure. Guided by these questions, we then present the results.}

\subsection{\revision{Experimental Design}}
\revision{Prior EmCom experiments have often explored pressures inspired by human evolution, such as ease of learning \cite{KIRBY2014108,NEURIPS2019_b0cf188d} among others \cite{vithanage,galke_raviv}, to encourage compositionality in emergent communication schemes. Meanwhile, prior work often assumes or favors concatenativity in \revision{evaluation} \cite{resnick,ueda2023on} despite the possibility of compositional but nonconcatenative communication schemes (\revision{e.g.,} our artificial and natural \revision{reference} languages).}

\revision{In order to achieve concatenative emergent languages with double articulation, we explore an \textit{ease of articulation} pressure. Human spoken languages are universally subject to phonological rules which determine the possibility of vocal sounds based on the environment and vocal tract physiology \cite{DEDIU20179}. The interface between phonology and morphology is a key area of study in linguistics, and a computational evolutionary model of morphology (as we hope to lay the groundwork for) would be incomplete without considering that morphemes are made up of meaningless components subject to similar rules independent of their meanings. Specifically, there are disallowed clusters of sounds in spoken languages (e.g., Hawaiian disallows consonant clusters), which may represent a significant pressure towards concatenative languages.
To test this, we added a term to the training loss for a toy simulated phonology for agents, where even-valued characters can not occur next to even-valued characters, and odd-valued characters can not occur next to odd-valued characters.\footnote{More details about this pressure provided in Appendix~\ref{apx:articulation details}.}}

In \revision{all} experiments, agents are trained with 8 different random initializations. \revision{ Learned languages are considered successful if the agent achieves 90\% accuracy on the task, which occurs in most cases.}\footnoteref{fn:agents}



    

\begin{table*}[ht]
    \centering
    \footnotesize
    \setlength{\tabcolsep}{4pt}
    \begin{tabular}{lccccccc}
        \toprule
        \textbf{Setting} &   \textbf{Art.}    & \textbf{TopSim}&\textbf{BoSDis}& \textbf{HAS BoSDis$^V_C$} & \textbf{BPE$_{96}$ BoSDis$^V_C$} & \textbf{BPELen$_{96}$}  & \textbf{F-TopSim $\delta$} \\\cmidrule(lr){1-2}\cmidrule(lr){3-6}\cmidrule(lr){7-7}\cmidrule(lr){8-8}
        $16\times 16\times 16$ &  \xmark   & .339$\pm$.034&.213$\pm$.047& 3.25$\pm$4.31& 1.25$\pm$.359& 3.82$\pm$.235& -.0418$\pm$.0210\\
        $16\times 16\times 16$ &  \cmark   & .330$\pm$.028&.196$\pm$.076& 5.98$\pm$8.53& 1.04$\pm$.359& 3.534$\pm$.133& -.0691$\pm$.0350\\\cmidrule(lr){1-2}\cmidrule(lr){3-6}\cmidrule(lr){7-7}\cmidrule(lr){8-8}
        $42\times 2\times 3$ &  \xmark   & .247$\pm$.119&.504$\pm$.0629& 1.79$\pm$.0943& 1.23$\pm$.0604& 3.257$\pm$ .225& -.0078$\pm$.028\\
        $42\times 2\times 3$ &  \cmark   & .144$\pm$ 0.102&.509$\pm$.0861& 1.42$\pm$1.03& 1.56$\pm$ 1.79& 2.34$\pm$.182& -.0083$\pm$.0178\\\cmidrule(lr){1-2}\cmidrule(lr){3-6}\cmidrule(lr){7-7}\cmidrule(lr){8-8}
 {$42 \times 2 \times 2 \times 2$}   &\xmark  & .0859 $\pm$.039&.6585 $\pm$ 0.593& 11.9$\pm$6.07& 7.23$\pm$ 2.82& 2.69 $\pm$ 0.229&.0415$\pm$.0289\\
 {$42 \times 2 \times 2 \times 2$}   &\cmark  & 0.146$\pm$.0860&.323$\pm$0.103& 8.79$\pm$ 10.7& 11.2 $\pm$ 8.04& 2.33$\pm$ .226&.0219$\pm$.0185\\\cmidrule(lr){1-2}\cmidrule(lr){3-6}\cmidrule(lr){7-7}\cmidrule(lr){8-8}
 {$42\times 6$} &  \xmark  & .0880$\pm$.0367&.582$\pm$.136& 2.17$\pm$ 6.07& 1.47$\pm$2.82& 2.69$\pm$0.212&N/A\\
 {$42\times 6$} &  \cmark  & .239 $\pm$.0645&.645$\pm$ .452& 2.16$\pm$ .775& 2.51$\pm$ 1.25& 2.33$\pm$0.181&N/A\\\bottomrule
    \end{tabular}
    \caption{\revision{Evaluation of compositionality, concatenativity, and degree of fusion of emergent languages in default and inflection Attr-Val settings, without and with articulation pressure (Art.). }}\label{tab:all results}

    \vspace{-10pt}
    
\end{table*}

\begin{figure*}
    \centering
    \includegraphics[width=1\linewidth]{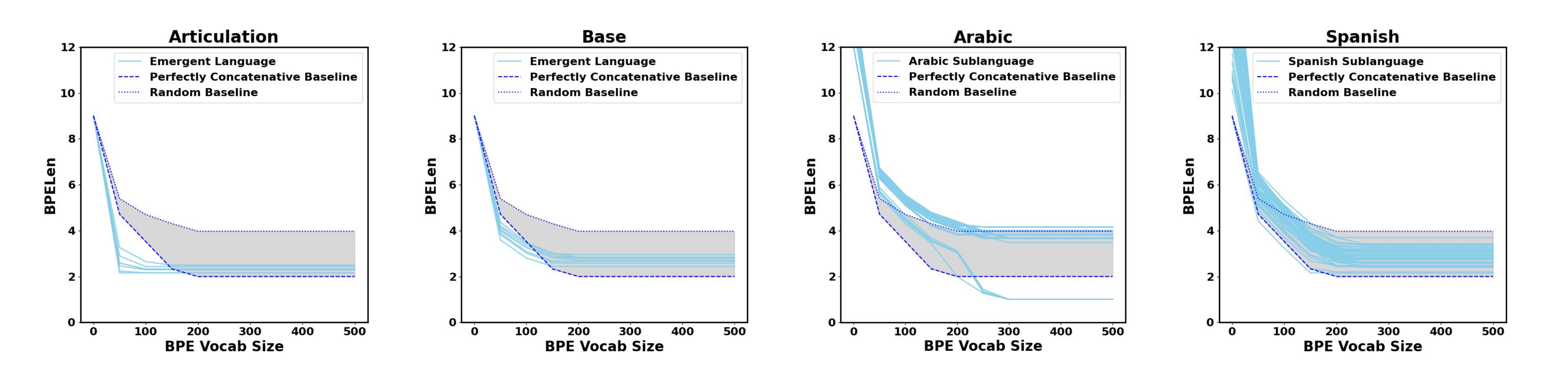}
    \vspace{-20pt}
    
    \caption{BPELen at varying $|V|$ for emergent and natural languages in the 42 $\times$ 3 inflection Attr-Val setting. \revision{Comprehensive plots for the full range of inflection experiments with natural language comparisons in Appendix~\ref{apx:more nat langs infl}.}}
    \label{fig:bpelen inflection}
\end{figure*}

\subsection{\revision{Experimental Results}}

\revision{In Table~\ref{tab:all results}, we summarize concatenativity and degree of fusion metrics in each Attr-Val setting, without and with articulation pressure. As shown there, BPE$_{96}$ BoSDis$^V_C$ > 1 on average across all emerged languages, implying the existence of multi-character meaningful segments and confirming that double articulation is achieved.}



\subsubsection{{Concatenativity and Articulation}}\label{sec:articulation}



\begin{figure}[ht!]
    \centering
    \includegraphics[width=1\linewidth]{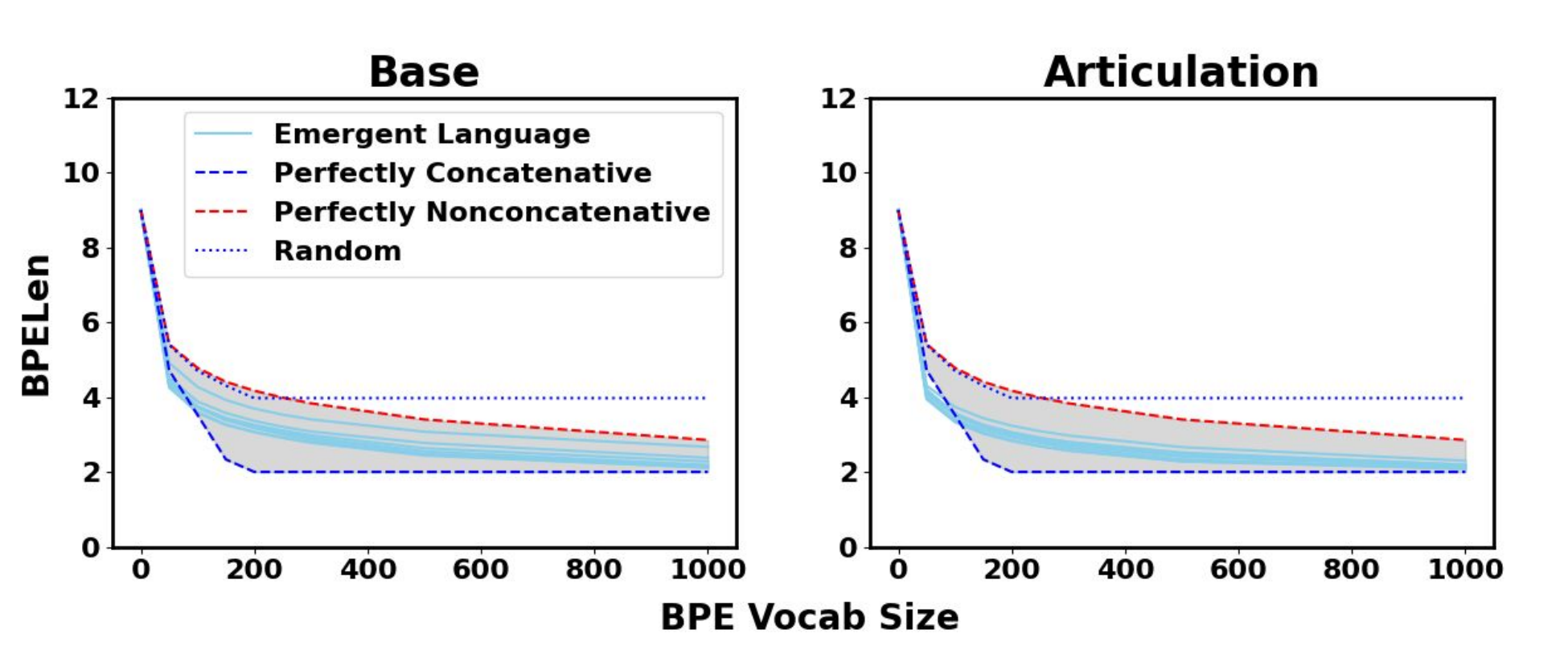}
    \vspace{-20pt}
    \caption{BPELen at varying $|V|$ for emergent languages in the default Attr-Val setting.}
    \vspace{-10pt}
    \label{fig:bpelen-graphs}
\end{figure}

\remove{Both without and with the ease of articulation pressure, }
We visualize the concatenativity (measured by BPELen) of emergent communication schemes in the default and inflection Attr-Val settings respectively in Figures~\ref{fig:bpelen inflection} and \ref{fig:bpelen-graphs}. 
First, we observe no inductive bias toward concatenativity in our RNN-based agents as previous works implicitly assume; instead, emergent languages \revision{fall} along a spectrum between perfectly concatenative and nonconcatenative. This may not be unrealistic: the natural languages also exhibit such a spectrum. Spanish appears relatively diverse, likely due to the presence of vowel patterns and auxilliary verbs. Arabic is more bimodal, with some very nonconcatenative sublanguages and some which have a high maximal compression due to being very fusional.  

However, the introduction of articulation pressure sharply decreases BPELen in emergent languages. We perform additional $t$-tests on the mean BPELen$_{96}$ of emergent languages between conditions, observing statistical significance in this change for both the default and inflection Attr-Val setting ($p\approx0.009$ and $p\approx 4.5\cdot10^{-7}$ respectively). This suggests ease of articulation is indeed a strong pressure towards concatenation, and the presence of ``unpronounceable'' clusters at the phonological (character) level is sufficient to induce concatenation at the morphological (symbol) level. More broadly, this may help explain the dominance of concatenative morphology in natural languages.

\subsubsection{{TopSim in Inflectional Languages}}
\begin{figure}
    \centering
    \includegraphics[width=0.8\linewidth]{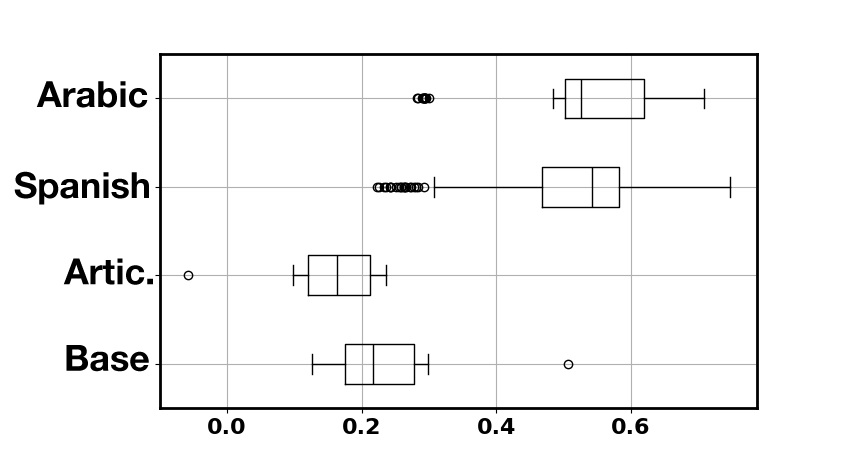}
    \vspace{-10pt}
    \caption{Topographic similarities of \revision{42 $\times$ 2 $\times$ 3 emergent (without and with articulation pressure) and natural languages. See Appendix \ref{apx:more nat langs infl} for additional conditions.}}
    \vspace{-10pt}
    \label{fig:infl_topsims}
\end{figure}

To shed more light on the compositionality of naturalistic communication schemes, we compare the TopSim of emergent and natural languages in the inflection Attr-Val setting in Figure~\ref{fig:infl_topsims}, including emergent languages with the articulation pressure.
Compared to natural languages, TopSim for emergent languages was low, except for one instance with a TopSim of .507 (in the range of natural languages). 
\revision{In the 42 $\times$ 2 $\times$ 2 $\times$ 2 and 42 $\times$ 6 settings, but not in the 42 $\times$ 2 $\times$ 3 setting, the articulation pressure resulted in a significant increase in TopSim. There seems to be a complex interplay between this constraint and the specific attribute-value setup, which we leave for future work to explore in more detail.}
Together, this suggests that emergent languages still have room for improvement in achieving humanlike topographic similarity, and there may remain more pressures toward this to identify and explore.

\subsubsection{\revision{Fusion} Under Inflection}

\remove{When reinterpreting Attr-Val as inflection, fusion becomes a naturalistic trait for languages.}

\remove{However, the average F-TopSim $\delta$ values are universally close to 0 in most emergent languages,\footnote{The only exception is the default setting without articulation pressure, where mean F-TopSim $\delta$ is positive, indicating a high degree of fusion. A $t$-test for mean F-TopSim $\delta$ with versus without articulation pressure in this setting } yields highly significant results ($p << 0.05$). A possible explanation for this is that fusion creates more total symbols, while articulation pressure discourages this, as there are only so many readily distinct legal symbols that abide by the articulation rule}. indicating an inconsistent degree of fusion.

\revision{As we would predict, the 16x16x16 setting was least fusional, with a quite negative F-TopSim $\delta$. In the 42 $\times$ 2 $\times$ 3 setting, F-TopSim $\delta$  was near 0, with 0 within a standard deviation of the mean. This implies there was some degree of fusion within these languages, since a completely non-fusional language would suffer a large drop in TopSim by fusing two attributes. The 42 $\times$ 2 $\times$ 2 $\times$ 2 setting was quite fusional, with a mean F-TopSim $\delta$ of 0.0415, i.e., by fusing the right attributes, we get a TopSim increase of over 0.04.)}

Suppletion (fusion of a root with grammatical features, e.g., \remove{in }\textit{went} as the past tense \remove{form }of \textit{go}) is much rarer than fusion \revision{between} \remove{two or more }grammatical features.
\remove{To judge whether this holds in EmCom, we identify the most fused attribute pair with the greatest F-TopSim $\delta$ in emergent and natural languages. }
\revision{Our EmCom results mirror this: the pair of lowest-cardinality attributes (representing tense and person)}
\revision{had the highest F-TopSim $\delta$ in 6 of 8 of the 42 $\times$ 2 $\times$ 3 emergent languages, suggesting these attributes are most likely to fuse. Similar results were observed for the 42 $\times$ 2 $\times$ 2 $\times$ 2 setting.}
\revision{This may be explained by symbolic complexity 
\cite{zhang-2025-combinatorial} increasing the least in fusing these attribute pairs; while 5 symbols are required to communicate the values for these two attributes before fusion, only 6 are required after.} Meanwhile, fusing the root attribute (with 42 values) with another attribute can drastically increase the symbolic complexity (i.e., 2-3 times).
This trend mirrors the fusion of fixed grammatical features in natural language\revision{, which we also observed in most of our generated Arabic and Spanish sublanguages.}
\remove{In Arabic, most sublanguages (78\%) also favored the fusion of these attributes. 
Unexpectedly, in Spanish, attributes 0 (root) and 1 (tense) fused most often (48\% of sublanguages), with (1,2) at 30\%. This may be due to the fact that specific roots can influence the form of inflectional affixes, e.g., the \textit{-amos} suffix only applies to \textit{-ar} verbs like \textit{llorar}; as such, these attributes are not represented by unique symbols, rather combinations of symbols.
Additionally, we observed that articulation pressure has a negative influence on degree of fusion across all experimental conditions.}

%% file: sections/6-conclusion.tex
\section{Conclusion}

In this work, we outline an experimental paradigm for emergent morphology, motivated by the principle of double articulation and a rich analogy between the Attr-Val game and inflectional morphology.
\remove{This environment has implicitly existed in a number of prior works, but ours is the first comprehensive treatment that specifically argues for its unique relationship to inflectional morphology.}
We review existing metrics for EmCom, proposing new ones for concatenativity 
\revision{and degree of fusion.}
\remove{In our empirical analyses, }We calibrate our metrics towards a wide array of idealized compositional schemes that are possible under this paradigm, as well as schemes based on natural language inflection. We then demonstrate through EmCom experiments the power of this reimagined setting to produce linguistically interesting analyses grounded in comparison to natural language. In particular, \remove{we show that }the double articulation aspect of our paradigm reveals simulated articulatory constraints as a promising pressure toward concatenativity, which prior work has often implicitly assumed in evaluations of compositionality and attempted to elicit through other pressures, and future work may use to test more specific morphophonological hypotheses. Further,
our emergent languages rederive the bias towards fusion of grammatical features in natural language.
\revision{Future work should adopt} \remove{the }small-vocabulary and inflectional Attr-Val settings for focused, linguistically motivated studies of morphology \revision{allowing direct comparison} to natural language.

%% file: sections/7-etc.tex
\section*{Limitations}

\paragraph{\revision{Naturalness of task inputs.}}
\revision{Like related works \cite{resnick,chaabouni-etal-2020-compositionality,ueda2023on}, we chose to focus on Attr-Val reconstruction as our EmCom task setting. This setting removes the requirement of agents to perceive key features from stimuli (e.g., images), instead only requiring agents to learn to communicate ground truth features, significantly simplifying the learning problem. However, we intentionally chose this setting to ensure comparability with \citet{ueda2023on}, and because our work is entirely focused on how EmCom agents learn to communicate features, thus possible perception errors would needlessly obstruct this effort. Further, considering that we aim to explore inflectional morphology in EmCom, the Attr-Val game is a natural analogy to existing models of morphology in linguistics, e.g., the lexeme-slot encoding of inflectional morphology \cite{wu-etal-2019-morphological}. Nonetheless, while out of scope for this work, an interesting direction for future work would be to investigate relationships between aspects of visual stimuli and morphology in emergent languages with small vocabularies.}

\paragraph{Representativeness of natural languages.}
In our natural language reference points, we only considered Arabic and Spanish, which are not a representative sample of the world's natural languages, and were chosen in part due to the availability of data. Without a much larger and more diverse dataset, we cannot definitively say whether a given emergent language is naturalistic. 
Additionally, the Attr-Val scheme we focused on is a rather small subset of the complexity of natural morphologies. 
A full analysis of a large set of natural languages, to cover this diversity of morphologies, is outside the scope of this paper, but we believe our examples will provide a solid initial point of comparison for emergent languages. \citet{boldt2024elccemergentlanguagecorpus} explored larger-scale cross comparisons of emergent languages, and we anticipate that the addition of parallel natural language data to such analyses would prove useful for future work.

\paragraph{Representativeness of articulation pressure.}
The choice of an even-odd toy phonological constraint for the articulation pressure was arbitrary, and not representative of the range of such constraints in natural language.
An alternative explanation for the increased concatenativity from this pressure is that the predictable even-odd alternation pattern resulting from this pressure simply made the data more compressible due to the fewer number of possible variations in messages. While this is only a simple proof of concept of the sorts of linguistically motivated experiments we can do when embracing morphology and double articulation in EmCom, we argue that phonological constraints in natural languages may provide a similar advantage in human language learning, and encourage future investigation of how they may influence concatenativity and other properties in EmCom.


%% file: sections/8-apx.tex
\appendix

\section{Agent Implementation Details and Task Accuracy}\label{apx:agent impl}
In line with prior related works, we implement the sender and receiver agents as single layer gated recurrent unit (GRU; \citealp{gru}) models using the EGG framework \cite{egg}, with hidden dimensions of 500. Observed states were implemented by concatenating one-hot vectors for each attribute.
Senders are optimized using REINFORCE \cite{reinforce} and receivers with backpropagation following \citet{NEURIPS2019_31ca0ca7} using the accuracy (i.e., proportion of attributes correctly reconstructed) as a loss, with cross-entropy loss as a supplementary loss term. It was found in initial experiments that this allowed smoother and more reliable convergence with lower sequence lengths, which is desirable for obtaining interpretable and minimally complex languages. Each run occurred on a single A40 GPU with 48GB VRAM and took about 8 hours, although many models converged before then. \revision{Models generally converged after about 400 training epochs.}

In the inflection Attr-Val setting, we additionally weighted the root accuracy so it represented 90\% of the accuracy optimized by the inflection-condition agents, for two reasons: initial experiments revealed unstable training without weighting, and this reflects that far more of the semantic content of a word is stored in the lexeme, rather than the slot, so the lexeme is more important to communicate. We also experimented with larger configurations with more ``roots,'' but we found that higher input sizes led to unstable training, where agents would learn either to inflect or distinguish roots, but not both. We attribute this to the increased symbolic complexity of the setting \cite{zhang-2025-combinatorial}. 

We ran experiments with agents 8 times for each setting with different random initializations. 
In the default Attr-Val setting, 7 of 8 emerged languages were successful, i.e., achieved greater than 90\% accuracy on the reconstruction task. 
For other experiments, including with the articulation pressure discussed in Section~\ref{sec:articulation} and the inflectional Attr-Val setting, all languages were considered successful. 

\section{\revision{Additional Notes On Inflection Setting}}\label{apx:inflection note}
\revision{In the inflection setting of Attr-Val, one possible limitation is the lack of control on the number of combinations of attribute values (rather than unique number of attribute values, which we control for instead). We conducted a preliminary experiment where we controlled for number of combinations rather than number of values, however these models universally failed to converge, instead learning to distinguish either grammatical features or roots, but never both.}

\revision{We believe that this failure to converge is related to the model capacity, as explored by \citet{resnick}, Specifically, we hypothesize that for languages with more unique values (i.e., symbolic complexity as defined in \citealp{zhang-2025-combinatorial}), the minimum model capacity for a successful language increases. While we could simply increase the size of this model (e.g., embedding and/or hidden size) to better encourage its convergence, it would no longer be comparable to other results. Meanwhile, if we increased the sizes of all models, this may influence compositionality in less complex configurations. As such, we decided that a comparison controlling for the number of combinations would not be possible within the scope of this paper.}

\revision{Nevertheless, we chose our specific Attr-Val configuration in the interest of comparability to natural language. To make a good comparison, we needed to be able to select features from natural language with equal or greater cardinalities to our artificial features. As such, there are only so many grammatical features we can introduce and find high-quality data for, and considering features with larger cardinalities would preclude a lot of natural language features like gender, number, person, etc. and severely limit our dataset. The specific values were chosen to allow this comparison and control for symbolic complexity.}

\section{Harris' Articulation Scheme}\label{apx:has mods}
\revision{We replicated the results in \citet{ueda2023on}
that C-TopSim is greater than W-Topsim. This result led the authors to suggest that Criterion 3 for Harris' Articulation Scheme (i.e., W-TopSim should be higher than C-TopSim) may be unsolvable.}
\begin{table}[b]
    \centering
    \begin{tabular}{c|c}
    \toprule
         C-TopSim&  W-TopSim\\
         \midrule
         0.611& 
    0.362\\
    \bottomrule
    \end{tabular}
    \caption{W-TopSim and C-TopSim for a perfectly concatenative language.}
    \label{tab:ueda followup}
\end{table}
\revision{As discussed in Section~\ref{sec:metrics}, we use ratios of BoSDis to judge whether segmented symbols are meaningful. We assert that this is a reasonable replacement for the intended purpose of the criterion, since it asks the question "do segments act as symbols which tend to refer to some meanings more than others."}

\citet{ueda2023on} report negative results on the basis of low W-TopSim, which does not hold for our replacement metric of the BosDis ratio.  Thus, we believe that emergent languages do exhibit HAS, and that HAS does not represent a significant gap between natural and emerged languages, as is claimed. The HAS algorithm, therefore represents a useful tool for interpreting emergent communication. 

\section{Positional Disentanglement}\label{apx:posdis}

Motivated by the possibility that symbols in different positions could indicate different attributes (common in natural languages), \citet{chaabouni-etal-2020-compositionality} propose positional disentanglement (\textbf{PosDis}): 

\begin{equation}
    \frac{1}{m}\sum_{v\in V}\frac{\mathcal{I}(s_j; a_v) - \mathcal{I}(s_j; b_v)}{\mathcal{H}(s_j)}
\end{equation}

For a communication scheme with messages of length $m$, PosDis calculates the average difference in mutual information $\mathcal{I}$ between the value of a symbol $s_j$ in position $j \in \lbrack 1, m \rbrack $ and attributes $a_v$ and $b_v$, which respectively have the most and second-most mutual information with $v$.\footnote{Given the reliance on a consistent message length, we only apply PosDis based on individual characters rather than symbols extracted from HAS and BPE.} This is then normalized by the entropy $\mathcal{H}$ of $s_j$.
PosDis will be high if attributes are uniquely distinguishable by positions of symbols in messages.
However, as we did not observe insightful trends with respect to this metric, we omitted PosDis from results presented in the paper to conserve space.

\section{Artificial Languages Extended Analysis}\label{apx:more art langs}
In this appendix, we provide extended discussions about concatenation and fusion in artificial languages. We then propose and evaluate additional nontrivial compositional artificial languages based on mutation and more general functional compositionality.

\paragraph{On Concatenation}
First, the level of concatenation in the first three artificial languages has minimal effect on TopSim and learnability, implying a lack of inductive bias toward concatenation in recurrent listener neural networks. 
Notably, this is not consistent with human results in linguistics \cite{PPR:PPR323827}, and implies additional pressures may be required to induce concatenative morphology.
Meanwhile, as expected, BoSDis$^V_C$ metrics indicate that both HAS and BPE successfully extract more meaningful morphemes from more concatenative languages, despite all artificial languages being compositional. In the mixed concatenative and variable length languages, HAS yields lower metrics, suggesting it may indeed be less effective in languages with randomness.
BPE can seemingly compensate for this as long as a sufficiently limited vocabulary size $|V|$ is chosen. The right vocabulary size is not obvious, however, as the BosDis$^V_C$ for BPE$_{\text{Max}}$ is relatively low for artificial languages, suggesting an over-segmented language that erases the disentanglement of symbols.
While segmentation algorithms behave as expected for systematic languages, these metrics are greater than 1 for random languages. This may result from over-sensitivity to noise in the absence of other signal, e.g., the presence of a few rare, highly predictive symbols may skew the metric upwards. This may also be attributed to a vanishingly small BoSDis for character-level segmentation. Future research should consider this edge case when comparing BoSDis before and after symbol segmentation.

\begin{figure}
    \centering
    \includegraphics[width=0.9\linewidth,trim={0 0 0 1.2cm},clip]{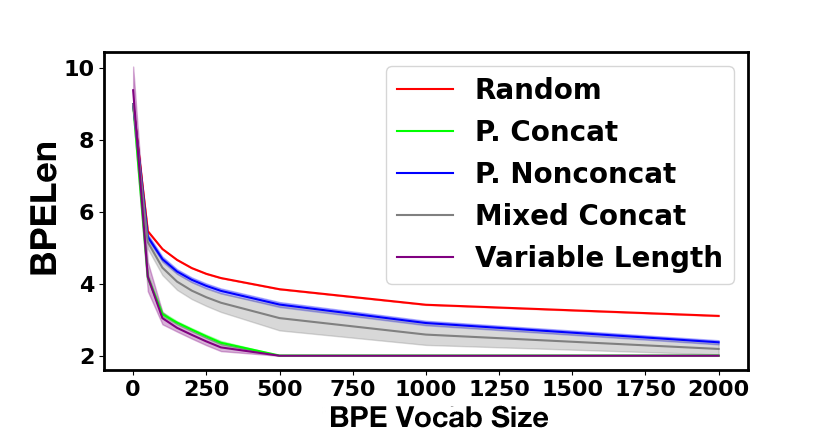}
    \vspace{-5pt}
    \caption{\revision{BPELen at varying $|V|$ for artificial languages, averaged for random seeds with $1\sigma$ confidence interval.}}
    \vspace{-12pt}
    \label{fig:artificial lang metrics}
\end{figure}

Focusing in on concatenativity as measured by HASLen and BPELen metrics, we unsurprisingly observe that the perfectly concatenative and variable length languages (which are both entirely concatenative) exhibit the highest concatenativity, except for BPELen$_\text{Max}$, which shows no sensitivity to the level of concatenativity in artificial languages. This further supports the possibility of over-segmenting languages into symbols. As such, in Figure~\ref{fig:artificial lang metrics}, we visualize the BPELen for a range of symbol vocabulary sizes. 
While minimal compression and maximal compression are similar across all artificial languages except the random language, there is a wide range in which BPELen behaves as expected. The mixed concatenative languages appear to fall on a spectrum between perfect concatenation and nonconcatenation, thus BPELen appears to be an effective measure of concatenativity. 
Meanwhile, due to the marked difference in BPELen between random and systematic languages, maximal compression may be an effective way to judge the amount of randomness in messages.

\paragraph{On Fusion}
F-TopSim $\delta$ values indicate that the fusion language is indeed the most fusional. We observe that F-TopSim $\delta$ for random languages is zero, which does not suggest \revision{fusion}, rather that TopSim does not change with respect to fusion of attributes (expected for a random language, which should not assign any meaningful symbols to any attribute values or pairs thereof). We observe that \revision{degree of fusion} does not vary significantly with respect to concatenativity, but increases in the variable length concatenative language.

\begin{figure}
    \centering
    \includegraphics[width=1\linewidth]{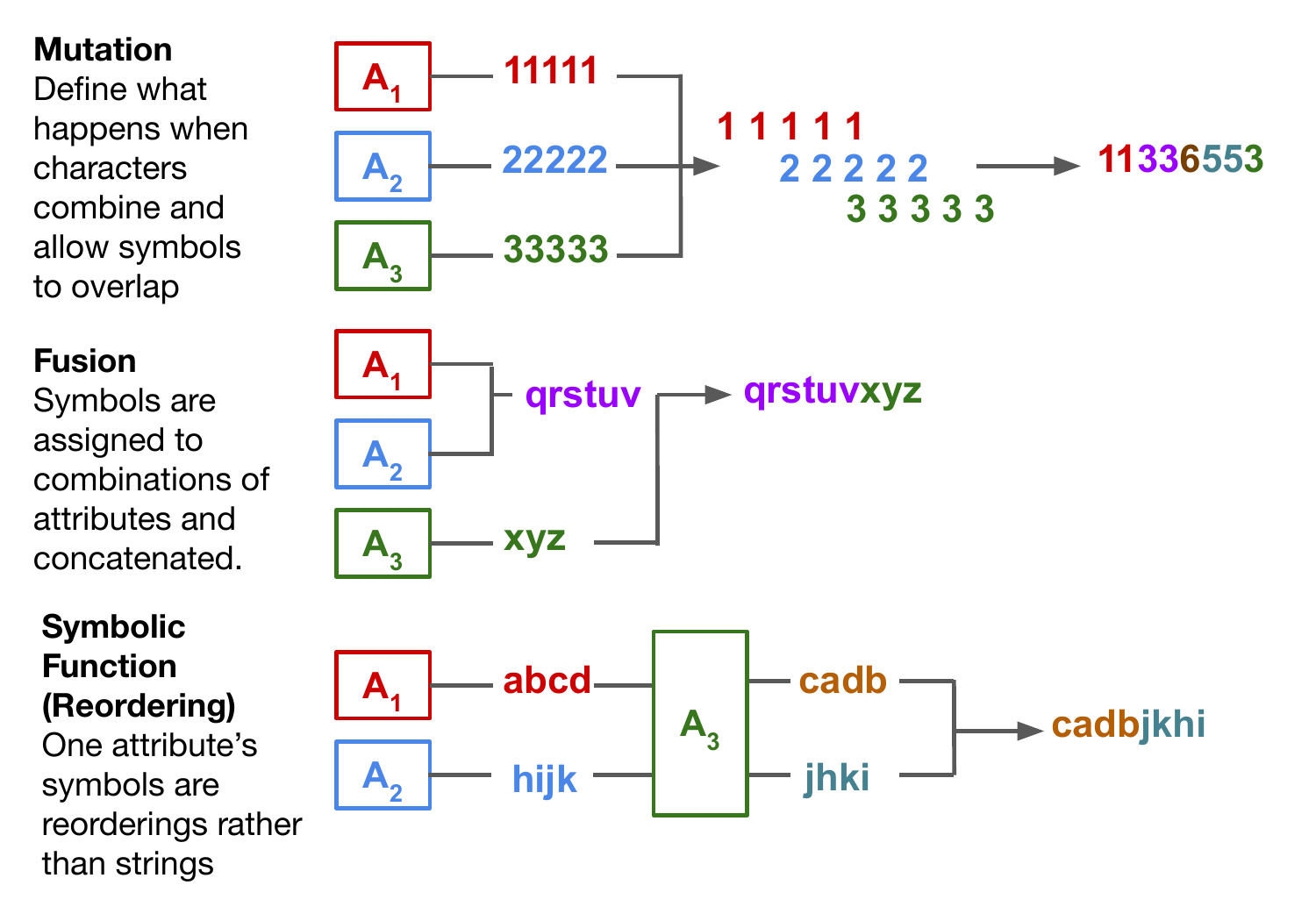}

    \vspace{-4pt}
    \caption{Artificial languages targeting mutation and symbolic functions (reordering). These languages define a unique symbol for each possible attribute value, and apply nontrivial composition operation for symbols.}
    \label{fig:artificial_lang 2}
\end{figure}

\paragraph{On Mutation and Functional Compositionality}
Traditional metrics for compositionality such as TopSim do not account for some types of nontrivial compositionality, such as phonemic mutation (e.g., of vowels in inflection of \textit{sing}, \textit{sang}, and \textit{sung}), or more general function-based compositionality, where symbols from some attributes serve as functions applied to other attributes. As visualized in Figure~\ref{fig:artificial_lang 2}, we defined additional languages for these phenomena. Specifically, following \citet{korbak2020measuring}, the \textit{mutation} language allows symbols to overlap, and combines overlapping characters' values by addition modulo $|C|$. We specifically consider two forms of the mutation language: Mutation$_3$, which consists of length 5 symbols each overlapping each other at 3 character positions, and Mutation$_0$, which consists of length 9 symbols each entirely overlapping. Lastly, the \textit{symbolic function} language defines one symbol as a function, specifically a \textit{reordering} function applied to the characters of another symbol. Beyond these artificial languages, we also consider a baseline \textit{random language}.

\revision{
\paragraph{On BoSDis$^V_C$ in Random Languages}
One may note that the BoSDis$^V_C$ for random languages is quite high. We expect this for 2 reasons:
\begin{enumerate}
    \item They have character-level BoSDis close to 0.
    \item By sheer random chance, some segments of characters will be better predictors of attributes than others. This results in a slight increase in symbol-level BoSDis. This results in a very high ratio, since even a small absolute increase may be 10x as great as the original BosDis. For this reason we don’t advocate putting much stock in the magnitude of the ratio. 
    \end{enumerate}
Whether this is a problem for the metric is a matter of interpretation. We believe that since random languages are very unique when all metrics, especially TopSim, are taken into account, the risk of misinterpreting a random language as a meaningfully segmented one is low. Furthermore, this enhances the confidence we can place in low BosDis ratio being an indicator of non-concatenativity, as these languages are qualitatively different from both random and concatenative languages (see BoSDiS$^V_C$< 1 for perfectly nonconcatenative languages in Table 1).
}

The evaluation metrics for these artificial languages are listed in Tables~\ref{tab:artificial language metrics apx} and \ref{tab:artificial language metrics apx 2}. 
Despite being compositional, these languages drastically impact evaluation metrics. The reordering and Mutation$_3$ language are substantially less compositional (according to TopSim and BoSDis), segmentable, and learnable than fully concatenative languages. In Mutation$_0$, where all characters are mutated twice, the languages were essentially indistinguishable from random languages by almost all metrics, including learnability. 
Even though this language consists entirely of messages formed by composing symbols according to simple, defined rules, in practice this compositionality may not matter or be recognized through existing metrics. 
The only metrics where there was a noticeable difference are the BoSDis$^V_C$ ratios for HAS and BPE, but the interpretations of these are unclear, since segments cannot be disentangled symbols for this type of language.

\begin{table*}[t]
    \centering
    \footnotesize
    \setlength{\tabcolsep}{4.3pt}
    \begin{tabular}{rccccccc} 
        \toprule
         \textbf{Language} &   \textbf{TopSim}&\textbf{BoSDis}&  \textbf{HAS BoSDis$^V_C$}&\textbf{BPE$_{96}$ BoSDis$^V_C$}  &\textbf{BPE$_{\text{Max}}$ BoSDis$^V_C$}  &\textbf{F-TopSim}&\textbf{Epochs}\\
         \cmidrule(lr){1-1}\cmidrule(lr){2-6}\cmidrule(lr){7-7}\cmidrule(lr){8-8}
          
 Reordering &    .227$\pm .01$&.042$\pm .03$&  23.24$\pm 22.30$&88.64$\pm 76.56$&13.44 $\pm$11.84&-0.006$\pm$0.01&20.25$\pm 5.47$\\
 Mutation$_3$ &    .359$\pm .01$&.060$\pm .00$&  40.37$\pm 0$&17.31$\pm .00$  &.578$\pm.25$&-.023$\pm$.01&24.88$\pm $9.85\\
 Mutation$_0$ &    .000$\pm .00$&.001$\pm .00$&  3146$\pm 1736$&7461$\pm 3091$  &17.204$\pm$5.46&.001$\pm$.00&100+\\
 
 \bottomrule
 \end{tabular}
    \normalsize
    \caption{Evaluation metrics for compositionality, \revision{degree of fusion}, and learnability of \revision{additional} artificial languages averaged over 8 random seeds. Learnability is measured by the average number of epochs that a listener agent takes to learn the language, i.e., exceed 99\% reconstruction accuracy. 1$\sigma$ confidence intervals are provided to within 2 decimal points, or whole numbers for large BoSDis$^V_C$ values. }
    \label{tab:artificial language metrics apx}
\end{table*}

\begin{table}[t]
    \centering
    \footnotesize
    \setlength{\tabcolsep}{5.45pt}
    \begin{tabular}{rccc} 
        \toprule
         \textbf{Language} &\textbf{HASLen}&\textbf{BPELen$_{96}$}&  \textbf{BPELen$_{\text{Max}}$} \\
         \cmidrule(lr){1-1}\cmidrule(lr){2-4}
          
 Reordering &2.268$\pm .17$&4.127$\pm.04$&  2.086$\pm .01$\\
 Mutation$_3$ &4.843$\pm .17$&4.461$\pm.04$&  2.273$\pm .05$\\
 Mutation$_0$ &1.560$\pm .38$&4.998$\pm.03$& 3.107$\pm .02$\\
 
 \bottomrule
 \end{tabular}
    \normalsize
    \caption{Evaluation metrics for concatenativity of artificial languages, averaged over 8 random seeds. 1$\sigma$ confidence intervals are provided to within 2 decimal points, or whole numbers for large BoSDis$^V_C$ values.}
    \label{tab:artificial language metrics apx 2}
\end{table}

\section{HASLen for Natural and Emergent Languages}\label{apx:haslen}

\begin{figure*}
    \centering
    \includegraphics[width=1\linewidth,trim={0 34cm 0 0cm},clip]{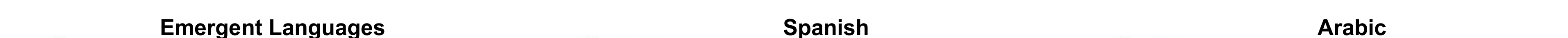}
    \includegraphics[width=1\linewidth,trim={0 0 0 17.5cm},clip]{inflect_figure.pdf}

\vspace{-20pt}

        \caption{HASLen of emergent and natural languages with respect to HAS entropy threshold.}
    \label{fig:infl_concat}
\end{figure*}

In Figure~\ref{fig:infl_concat}, we visualize the HASLen for emergent languages (on average) along with Spanish and Arabic natural language reference points. Following observations in Section~\ref{sec:artificial lang}, we observe that Spanish and Arabic are scored with a higher HASLen than random languages. This further demonstrates HAS' sensitivity to randomness, making it difficult to use for reference points of concatenativity (despite its effectiveness in identifying symbol boundaries).

\section{Segmentation Analysis of Natural Languages}\label{apx:segment meaningfulness}
As shown in Table~\ref{tab:sp_ar_b}, when calculating the BoSDis$^V_C$ ratio for these languages, we find that up to 76\% of generated Spanish sublanguages\footnote{Some failures here could be attributed to nonconcatenativity in the dataset caused by prepended auxiliary verbs.} can be meaningfully segmented by either HAS or BPE (i.e., ratio greater than 1), while virtually none of the Arabic sublanguages can be. On the latter, this demonstrates how existing tools for symbol segmentation are insufficient to capture highly nonconcatenative morphologies.

\begin{table}
    \centering
    
    \begin{tabular}{llc}
    \toprule
          & HAS&BPE$_{96}$\\
         \midrule
          Spanish & 0.76 &0.74\\
 Arabic&  0.08&0.00\\
 \bottomrule
 \end{tabular}
    \caption{Proportion of sublanguages meaningfully segmented (BoSDis$^V_C$>1) by BPE$_{96}$ and HAS.}
    \label{tab:sp_ar_b}
\end{table}

\section{Ease-of-Learning Pressure Experiment}\label{apx:iterated learning}

\begin{figure}[ht]
    \centering
    \includegraphics[width=1\linewidth,trim={17cm 0 17cm 0},clip]{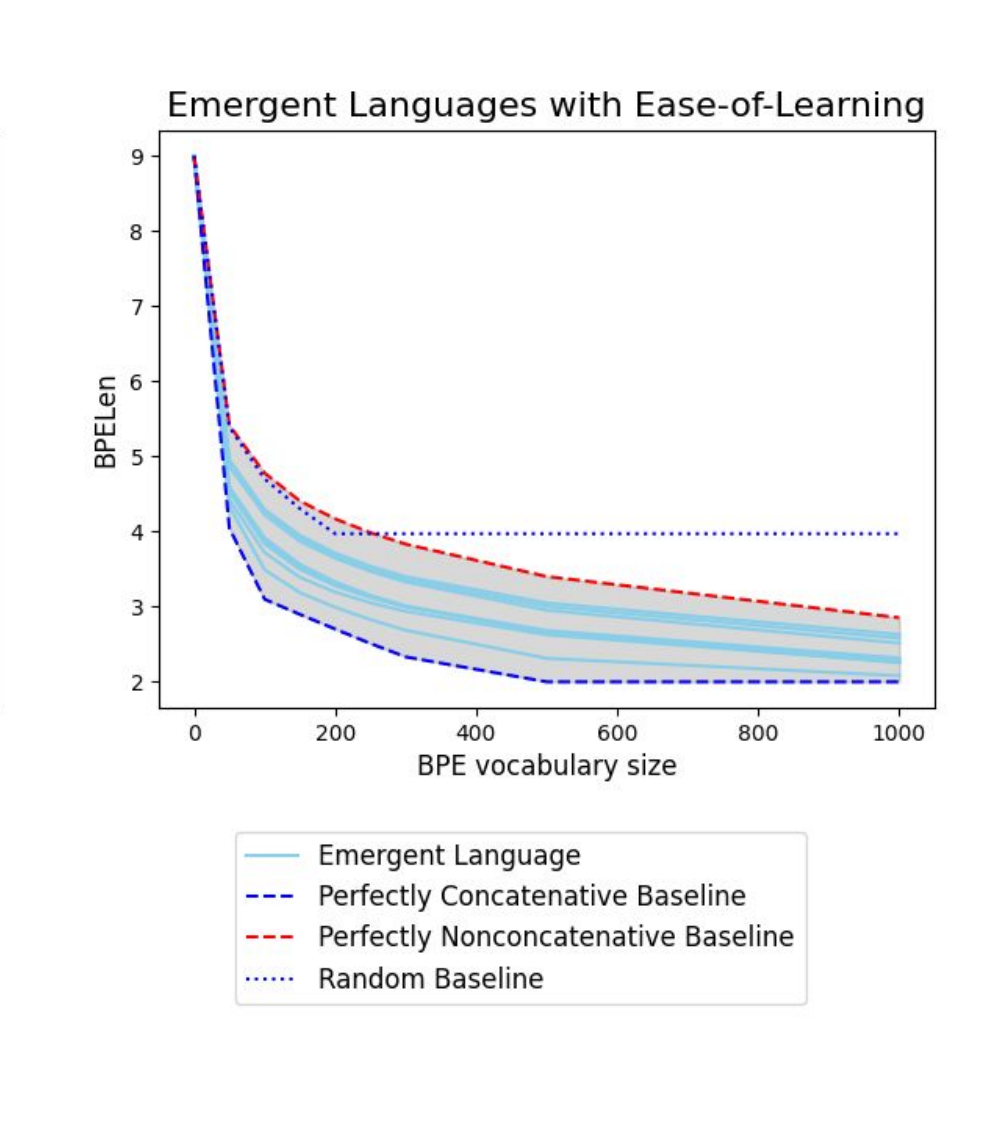}
    \caption{BPELen (concatenativty) of emerged languages with ease-of-learning pressure at varying symbol vocabulary sizes $|V|$.}
    \label{fig:bpelen-graphs-iter}
\end{figure}

Inspired by prior work in iterated learning \cite{KIRBY2014108}, we implement an experiment incentivizing ease of learning in emergent languages \cite{NEURIPS2019_b0cf188d}. Specifically, every 100 epochs, we reset the listener's network weights. In theory, this should encourage emergent languages that are easier for the listener to learn quickly, which \citeauthor{NEURIPS2019_b0cf188d} show improves compositionality of the resulting languages. To understand whether this ease of learning is attributed to concatenativity, which prior work has typically not distinguished from compositionality, Figure~\ref{fig:bpelen-graphs-iter} visualizes the BPELen of emergent languages with this pressure in the default Attr-Val setting (i.e., 3 attributes each with 16 values).
Interestingly, we find that the languages resulting from this pressure are not significantly more concatenative than those without it shown in Figure~\ref{fig:bpelen-graphs}. Such a result is unexpected from an intuitive standpoint, but aligns with the results in \ref{fig:artificial lang metrics}, which show that concatenativity, by itself, does not have a relationship with learnability. This motivates future research around how various pressures may influence fine-grained properties of emergent languages, and how much of human language structure can be explained by the generational (iterated) learning paradigm. 

\section{Articulation Pressure Details}\label{apx:articulation details}

Formally, for a message $s$ comprised of a sequence of integer tokens, the loss term for ease of articulation pressure is computed as:

\begin{multline}
    a(s_i, s_j)=
    \begin{cases}
    1 & s_i \equiv s_j\mod2\\
    0 & otherwise
    \end{cases}
    \\\mathcal{L}_{articulation} =\epsilon\sum_{i=1}^{|s|-1}a(s_i, s_{i+1})
\end{multline}
$\epsilon$ is a hyperparameter which controls how 'strict' the phonological rule is. We tried 1, 10, and 100 as values and found that 10 resulted in fairly strict adherence without incurring optimization problems, so $\epsilon=10$  in our reported experiment. 

8 of 8 seeds in the default setting succeeded, and 7 of 8 of the inflectional setting. 
In the default Attr-Val setting, all of the successful articulatory languages exhibited $\text{BoSDis}_C^V > 1$. 
In the inflection setting, 3 of 7 of the successful languages exhibited $\text{BoSDis}_C^V > 1$, indicating that not all of the segmentations were meaningful. We hypothesize that the additional constraint caused difficulties in segmenting, or alternatively that the pressure may have also resulted in highly nonconcatenative languages that appear concatenative due to the additional pattern. The fact that the intersection of these two conditions results in a somewhat unintuitive result underscores the need for future work to understand the fine-grained details of EmCom.

\section{\revision{Additional Natural Language Comparisons}}\label{apx:more nat langs infl}
\begin{figure*}
    \centering
    \includegraphics[width=\linewidth]{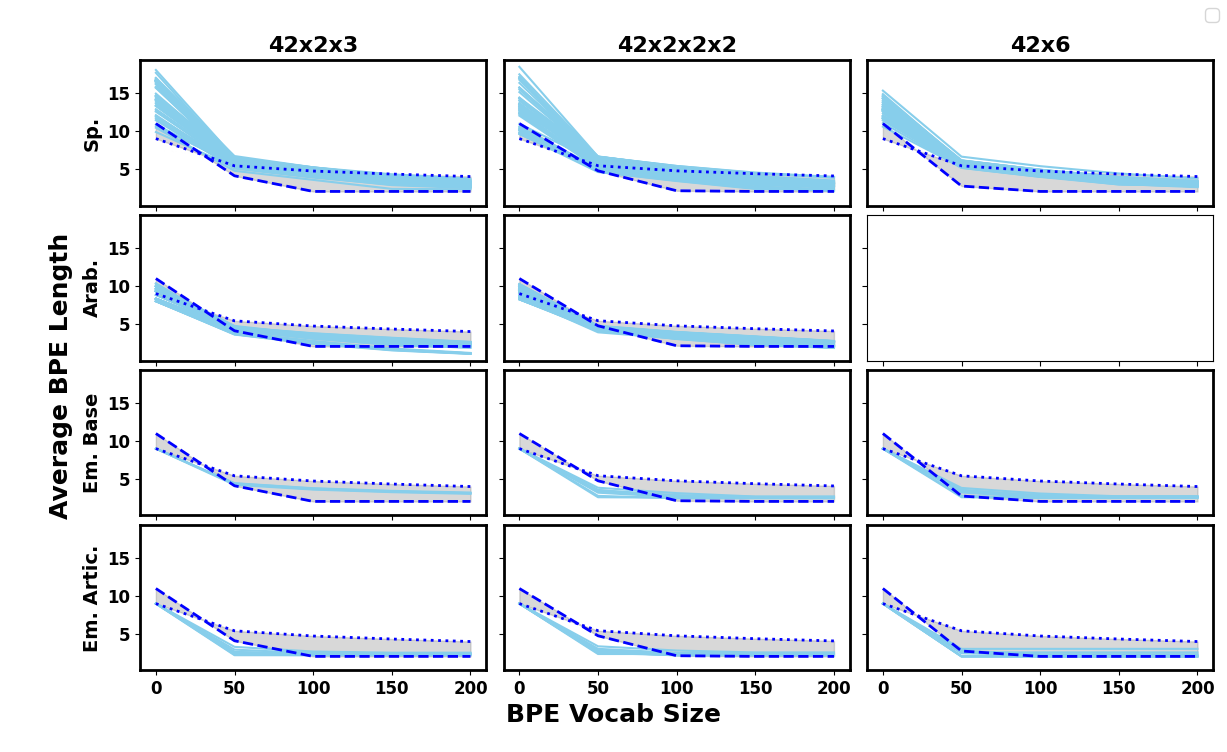}
    \caption{\revision{BPE compression for natural languages (Spanish and Arabic) and emergent languages across all inflection conditions. Dotted blue lines: random baselines; dashed blue lines: perfectly concatenative baselines; solid light blue lines: sublanguages or emergent languages, depending on condition.}}
    \label{fig:bpe_grid}
\end{figure*}

\begin{figure*}
    \centering
    \includegraphics[width=1\linewidth]{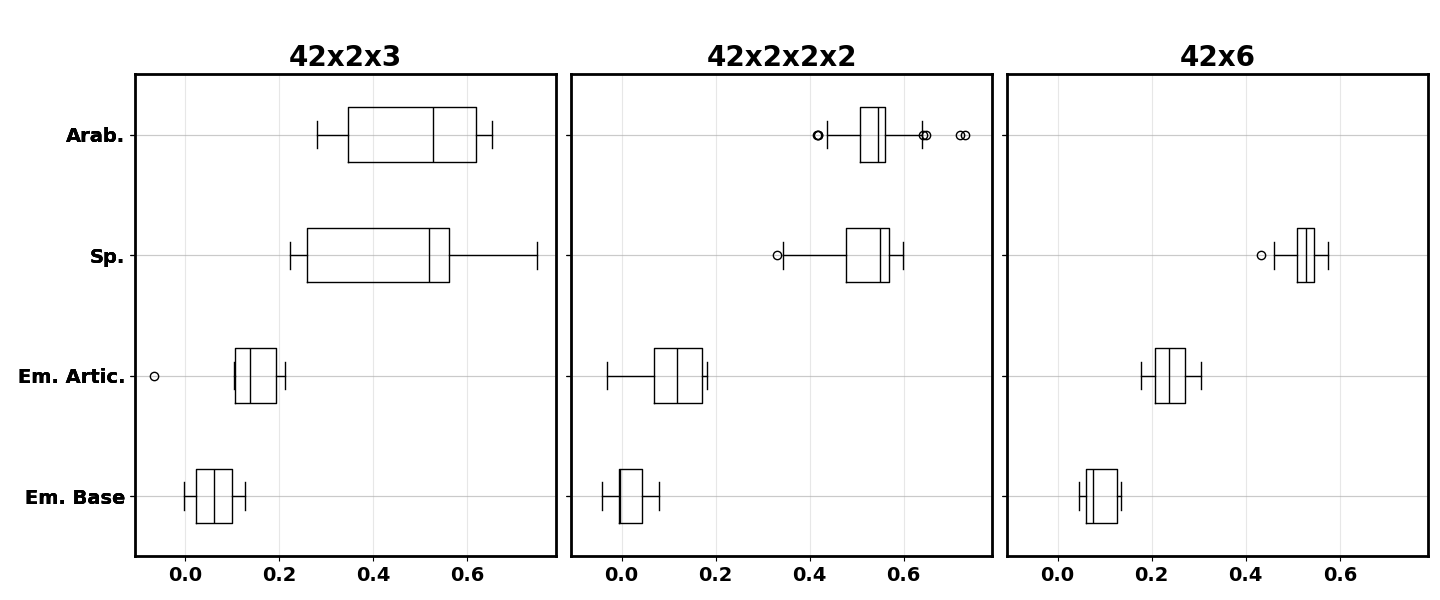}
    \caption{\revision{TopSim for natural languages (Spanish and Arabic) and emergent languages across all inflection conditions.}}
    \label{fig:topsim_grid}
\end{figure*}

\revision{Figure~\ref{fig:bpe_grid} shows BPE compression computed for Spanish, Arabic, emergent languages without articulation pressure (Em. Base) and emergent languages with articulation pressure (Em. Artic.). Figure~\ref{fig:topsim_grid} shows TopSim ranges for the same experiments. Due to lack of a sufficiently large grammatical feature in our Arabic dataset, we were unable to compute an Arabic baseline for the 42 $\times$ 6 condition.} 


\section{License Information}

We provide license information and links for the following artifacts used in this work:

\begin{itemize}
    \item EGG Framework: MIT License. 
    \subitem \url{https://github.com/facebookresearch/EGG}
    \subitem License: \url{https://github.com/facebookresearch/EGG/blob/main/LICENSE}
    \item Fred Jehle's Spanish Verb Dataset: Creative Commons Attribution-NonCommercial-ShareAlike 3.0 Unported License. 
    \subitem \url{https://github.com/ghidinelli/fred-jehle-spanish-verbs}
    \subitem License: \url{https://creativecommons.org/licenses/by-nc-sa/3.0/}
    \item Arabic-Urdu Conjugation Dataset 
    \subitem \url{https://github.com/haqnawaz99/Arabic-Urdu-Conjugation-Dataset}
\end{itemize}

%% file: acl_latex.bbl
\begin{thebibliography}{48}
\providecommand{\natexlab}[1]{#1}

\bibitem[{Archangeli and Pulleyblank(2016)}]{archangeli2016emergent}
Diana Archangeli and Douglas Pulleyblank. 2016.
\newblock Emergent morphology.
\newblock In \emph{Morphological metatheory}, pages 237--270. John Benjamins Publishing Company.

\bibitem[{Boldt and Mortensen(2024{\natexlab{a}})}]{boldt2024elccemergentlanguagecorpus}
Brendon Boldt and David Mortensen. 2024{\natexlab{a}}.
\newblock \href {https://arxiv.org/abs/2407.04158} {Elcc: the emergent language corpus collection}.
\newblock \emph{Preprint}, arXiv:2407.04158.

\bibitem[{Boldt and Mortensen(2024{\natexlab{b}})}]{boldt2024a}
Brendon Boldt and David~R Mortensen. 2024{\natexlab{b}}.
\newblock \href {https://openreview.net/forum?id=jesKcQxQ7j} {A review of the applications of deep learning-based emergent communication}.
\newblock \emph{Transactions on Machine Learning Research}.

\bibitem[{Brighton and Kirby(2006)}]{topsim}
Henry Brighton and Simon Kirby. 2006.
\newblock \href {https://doi.org/10.1162/artl.2006.12.2.229} {Understanding linguistic evolution by visualizing the emergence of topographic mappings}.
\newblock \emph{Artificial Life}, 12:229--242.

\bibitem[{Chaabouni et~al.(2020)Chaabouni, Kharitonov, Bouchacourt, Dupoux, and Baroni}]{chaabouni-etal-2020-compositionality}
Rahma Chaabouni, Eugene Kharitonov, Diane Bouchacourt, Emmanuel Dupoux, and Marco Baroni. 2020.
\newblock \href {https://doi.org/10.18653/v1/2020.acl-main.407} {Compositionality and generalization in emergent languages}.
\newblock In \emph{Proceedings of the 58th Annual Meeting of the Association for Computational Linguistics}, pages 4427--4442, Online. Association for Computational Linguistics.

\bibitem[{Chaabouni et~al.(2019)Chaabouni, Kharitonov, Dupoux, and Baroni}]{NEURIPS2019_31ca0ca7}
Rahma Chaabouni, Eugene Kharitonov, Emmanuel Dupoux, and Marco Baroni. 2019.
\newblock \href {https://proceedings.neurips.cc/paper_files/paper/2019/file/31ca0ca71184bbdb3de7b20a51e88e90-Paper.pdf} {Anti-efficient encoding in emergent communication}.
\newblock In \emph{Advances in Neural Information Processing Systems}, volume~32. Curran Associates, Inc.

\bibitem[{Chaabouni et~al.(2022)Chaabouni, Strub, Altch{\'e}, Tarassov, Tallec, Davoodi, Mathewson, Tieleman, Lazaridou, and Piot}]{chaabouni2022emergent}
Rahma Chaabouni, Florian Strub, Florent Altch{\'e}, Eugene Tarassov, Corentin Tallec, Elnaz Davoodi, Kory~Wallace Mathewson, Olivier Tieleman, Angeliki Lazaridou, and Bilal Piot. 2022.
\newblock \href {https://openreview.net/forum?id=AUGBfDIV9rL} {Emergent communication at scale}.
\newblock In \emph{International Conference on Learning Representations}.

\bibitem[{Cho et~al.(2014)Cho, van Merrienboer, Gulcehre, Bahdanau, Bougares, Schwenk, and Bengio}]{gru}
Kyunghyun Cho, Bart van Merrienboer, Caglar Gulcehre, Dzmitry Bahdanau, Fethi Bougares, Holger Schwenk, and Yoshua Bengio. 2014.
\newblock \href {https://arxiv.org/abs/1406.1078} {Learning phrase representations using rnn encoder-decoder for statistical machine translation}.
\newblock In \emph{Proceedings of the 2014 Conference on Empirical Methods in Natural Language Processing,}, page 1724–1734.

\bibitem[{Conklin and Smith(2023)}]{conklin2023compositionality}
Henry Conklin and Kenny Smith. 2023.
\newblock \href {https://openreview.net/forum?id=-Yzz6vlX7V-} {Compositionality with variation reliably emerges in neural networks}.
\newblock In \emph{The Eleventh International Conference on Learning Representations}.

\bibitem[{Dediu et~al.(2017)Dediu, Janssen, and Moisik}]{DEDIU20179}
Dan Dediu, Rick Janssen, and Scott~R. Moisik. 2017.
\newblock \href {https://doi.org/10.1016/j.langcom.2016.10.002} {Language is not isolated from its wider environment: Vocal tract influences on the evolution of speech and language}.
\newblock \emph{Language \& Communication}, 54:9--20.
\newblock The multimodal origins of linguistic communication.

\bibitem[{Dekker(2024)}]{dekker}
Peter Dekker. 2024.
\newblock \emph{Identifying drivers of language change using agent-based models}.
\newblock Ph.D. thesis, Vrije Universiteit Brussel.

\bibitem[{Elsner et~al.(2020)Elsner, Johnson, Antetomaso, and Sims}]{elsner2020stop}
Micha Elsner, Martha~B Johnson, Stephanie Antetomaso, and Andrea~D Sims. 2020.
\newblock Stop the morphological cycle, i want to get off: Modeling the development of fusion.
\newblock \emph{Society for Computation in Linguistics}, 3(1).

\bibitem[{Finley and Newport(2021)}]{PPR:PPR323827}
Sara Finley and Elissa Newport. 2021.
\newblock \href {https://doi.org/10.31234/osf.io/xqy4k} {Segmentation of root and pattern morphology}.
\newblock \emph{PsyArXiv}.

\bibitem[{Fullwood(2018)}]{fullwood2018biases}
Michelle~Alison Fullwood. 2018.
\newblock \emph{Biases in segmenting non-concatenative morphology}.
\newblock Ph.D. thesis, Massachusetts Institute of Technology.

\bibitem[{Gage(1994)}]{bpe}
Philip Gage. 1994.
\newblock A new algorithm for data compression.
\newblock \emph{The C Users Journal}, 2:23--38.

\bibitem[{Galke et~al.(2022)Galke, Ram, and Raviv}]{galke2022emergentcommunicationunderstandinghuman}
Lukas Galke, Yoav Ram, and Limor Raviv. 2022.
\newblock \href {https://arxiv.org/abs/2204.10590} {Emergent communication for understanding human language evolution: What's missing?}
\newblock In \emph{EmeCom at ICLR 2022}.

\bibitem[{Galke and Raviv(2024)}]{galke_raviv}
Lukas Galke and Limor Raviv. 2024.
\newblock \href {https://doi.org/10.34842/3VR5-5R49} {Learning and communication pressures in neural networks: Lessons from emergent communication}.
\newblock \emph{Language Development Research}, 5.

\bibitem[{Gupta et~al.(2020)Gupta, Resnick, Foerster, Dai, and Cho}]{gupta-etal-2020-compositionality}
Abhinav Gupta, Cinjon Resnick, Jakob Foerster, Andrew Dai, and Kyunghyun Cho. 2020.
\newblock \href {https://doi.org/10.18653/v1/2020.repl4nlp-1.5} {Compositionality and capacity in emergent languages}.
\newblock In \emph{Proceedings of the 5th Workshop on Representation Learning for NLP}, pages 34--38, Online. Association for Computational Linguistics.

\bibitem[{Gutierrez-Vasques et~al.(2023)Gutierrez-Vasques, Bentz, and Samardžić}]{10.1162/coli_a_00489}
Ximena Gutierrez-Vasques, Christian Bentz, and Tanja Samardžić. 2023.
\newblock \href {https://doi.org/10.1162/coli_a_00489} {Languages through the looking glass of bpe compression}.
\newblock \emph{Computational Linguistics}, 49(4):943--1001.

\bibitem[{Gutierrez-Vasques et~al.(2021)Gutierrez-Vasques, Bentz, Sozinova, and Samardzic}]{gutierrez-vasques-etal-2021-characters}
Ximena Gutierrez-Vasques, Christian Bentz, Olga Sozinova, and Tanja Samardzic. 2021.
\newblock \href {https://doi.org/10.18653/v1/2021.eacl-main.302} {From characters to words: the turning point of {BPE} merges}.
\newblock In \emph{Proceedings of the 16th Conference of the European Chapter of the Association for Computational Linguistics: Main Volume}, pages 3454--3468, Online. Association for Computational Linguistics.

\bibitem[{Hockett(1960)}]{hockett}
Charles Hockett. 1960.
\newblock The origin of speech.
\newblock \emph{Scientific American}, 203:88--96.

\bibitem[{Jehle(2011)}]{jehle-verb-database}
Fred Jehle. 2011.
\newblock \href {https://github.com/ghidinelli/fred-jehle-spanish-verbs} {Fred {Jehle's} conjugated {Spanish} verb database}.

\bibitem[{Kharitonov et~al.(2019)Kharitonov, Chaabouni, Bouchacourt, and Baroni}]{egg}
Eugene Kharitonov, Rahma Chaabouni, Diane Bouchacourt, and Marco Baroni. 2019.
\newblock \href {https://doi.org/10.18653/v1/D19-3010} {Egg: a toolkit for research on emergence of language in games}.
\newblock In \emph{Proceedings of the 2019 Conference on Empirical Methods in Natural Language Processing and the 9th International Joint Conference on Natural Language Processing (EMNLP-IJCNLP): System Demonstrations}, page 55–60.

\bibitem[{Kirby et~al.(2014)Kirby, Griffiths, and Smith}]{KIRBY2014108}
Simon Kirby, Tom Griffiths, and Kenny Smith. 2014.
\newblock \href {https://doi.org/10.1016/j.conb.2014.07.014} {Iterated learning and the evolution of language}.
\newblock \emph{Current Opinion in Neurobiology}, 28:108--114.
\newblock SI: Communication and language.

\bibitem[{Korbak et~al.(2020)Korbak, Zubek, and Rączaszek-Leonardi}]{korbak2020measuring}
Tomasz Korbak, Julian Zubek, and Joanna Rączaszek-Leonardi. 2020.
\newblock \href {https://arxiv.org/abs/2010.15058} {Measuring non-trivial compositionality in emergent communication}.
\newblock \emph{Preprint}, arXiv:2010.15058.

\bibitem[{Kottur et~al.(2017)Kottur, Moura, Lee, and Batra}]{kottur-etal-2017-natural}
Satwik Kottur, José Moura, Stefan Lee, and Dhruv Batra. 2017.
\newblock \href {https://aclanthology.org/D17-1321/} {Natural language does not emerge ‘naturally’ in multi-agent dialog}.
\newblock In \emph{Proceedings of the 2017 Conference on Empirical Methods in Natural Language Processing}, page 2962–2967. Copenhagen, Denmark. Association for Computational Linguistics.

\bibitem[{Kouwenhoven et~al.(2024)Kouwenhoven, Peeperkorn, Van~Dijk, and Verhoef}]{kouwenhoven-etal-2024-curious}
Tom Kouwenhoven, Max Peeperkorn, Bram Van~Dijk, and Tessa Verhoef. 2024.
\newblock \href {https://doi.org/10.18653/v1/2024.cmcl-1.5} {The curious case of representational alignment: Unravelling visio-linguistic tasks in emergent communication}.
\newblock In \emph{Proceedings of the Workshop on Cognitive Modeling and Computational Linguistics}, pages 57--71, Bangkok, Thailand. Association for Computational Linguistics.

\bibitem[{Lazaridou et~al.(2018)Lazaridou, Hermann, Tuyls, and Clark}]{lazaridou2018emergence}
Angeliki Lazaridou, Karl~Moritz Hermann, Karl Tuyls, and Stephen Clark. 2018.
\newblock \href {https://openreview.net/forum?id=HJGv1Z-AW} {Emergence of linguistic communication from referential games with symbolic and pixel input}.
\newblock In \emph{International Conference on Learning Representations}.

\bibitem[{Levenshtein(1965)}]{Levenshtein1965BinaryCC}
Vladimir~I. Levenshtein. 1965.
\newblock \href {https://api.semanticscholar.org/CorpusID:60827152} {Binary codes capable of correcting deletions, insertions, and reversals}.
\newblock \emph{Soviet physics. Doklady}, 10:707--710.

\bibitem[{Lewis(1979)}]{lewis-signalling-game}
David Lewis. 1979.
\newblock \href {https://doi.org/10.1007/BF00258436} {Scorekeeping in a language game}.
\newblock \emph{Journal of Philosophical Logic}, 8:339–359.

\bibitem[{Li and Bowling(2019)}]{NEURIPS2019_b0cf188d}
Fushan Li and Michael Bowling. 2019.
\newblock \href {https://proceedings.neurips.cc/paper_files/paper/2019/file/b0cf188d74589db9b23d5d277238a929-Paper.pdf} {Ease-of-teaching and language structure from emergent communication}.
\newblock In \emph{Advances in Neural Information Processing Systems}, volume~32. Curran Associates, Inc.

\bibitem[{Lian et~al.(2023)Lian, Bisazza, and Verhoef}]{lian-etal-2023-communication}
Yuchen Lian, Arianna Bisazza, and Tessa Verhoef. 2023.
\newblock \href {https://doi.org/10.1162/tacl_a_00587} {Communication drives the emergence of language universals in neural agents: Evidence from the word-order/case-marking trade-off}.
\newblock \emph{Transactions of the Association for Computational Linguistics}, 11:1033--1047.

\bibitem[{Lian et~al.(2024)Lian, Verhoef, and Bisazza}]{lian-etal-2024-nellcom}
Yuchen Lian, Tessa Verhoef, and Arianna Bisazza. 2024.
\newblock \href {https://doi.org/10.18653/v1/2024.conll-1.19} {{N}e{LLC}om-{X}: A comprehensive neural-agent framework to simulate language learning and group communication}.
\newblock In \emph{Proceedings of the 28th Conference on Computational Natural Language Learning}, pages 243--258, Miami, FL, USA. Association for Computational Linguistics.

\bibitem[{McCarthy and Prince(1995)}]{faithfulness_identity}
John McCarthy and Alan Prince. 1995.
\newblock Faithfulness and reduplicative identity.
\newblock \emph{Papers in Optimality Theory. University of Massachusetts Occasional Papers}, 18.

\bibitem[{Nawaz et~al.(2025)Nawaz, Elobaid, Al-Laith, and Ullah}]{nawaz-etal-2025-automated}
Haq Nawaz, Manal Elobaid, Ali Al-Laith, and Saif Ullah. 2025.
\newblock \href {https://aclanthology.org/2025.abjadnlp-1.14/} {Automated generation of {A}rabic verb conjugations with multilingual {U}rdu translation: An {NLP} approach}.
\newblock In \emph{Proceedings of the 1st Workshop on NLP for Languages Using Arabic Script}, pages 136--143, Abu Dhabi, UAE. Association for Computational Linguistics.

\bibitem[{Nowak and Krakauer(1999)}]{nowak}
Martin Nowak and David Krakauer. 1999.
\newblock \href {https://www.pnas.org/doi/epdf/10.1073/pnas.96.14.8028} {The evolution of language}.
\newblock \emph{Proc. Natl. Acad. Sci. USA}, 96:8028–8033.

\bibitem[{Rathi et~al.(2021)Rathi, Hahn, and Futrell}]{rathi2021information}
Neil Rathi, Michael Hahn, and Richard Futrell. 2021.
\newblock An information-theoretic characterization of morphological fusion.
\newblock \emph{An Information-Theoretic Characterization of Morphological Fusion}.

\bibitem[{Rathi et~al.(2022)Rathi, Hahn, and Futrell}]{rathi2022explaining}
Neil Rathi, Michael Hahn, and Richard Futrell. 2022.
\newblock Explaining patterns of fusion in morphological paradigms using the memory--surprisal tradeoff.
\newblock In \emph{Proceedings of the annual meeting of the cognitive science society}, volume~44.

\bibitem[{Resnick et~al.(2020)Resnick, Gupta, Foerster, Dai, and Cho}]{resnick}
Cinjon Resnick, Abhinav Gupta, Jakob Foerster, Andrew Dai, and Kyunghyun Cho. 2020.
\newblock \href {https://arxiv.org/abs/1910.11424} {Capacity, bandwidth, and compositionality in emergent language learning}.
\newblock In \emph{Proceedings of the 19th International Conference on Autonomous Agents and Multiagent Systems}.

\bibitem[{Sennrich et~al.(2016)Sennrich, Haddow, and Birch}]{sennrich-etal-2016-neural}
Rico Sennrich, Barry Haddow, and Alexandra Birch. 2016.
\newblock \href {https://doi.org/10.18653/v1/P16-1162} {Neural machine translation of rare words with subword units}.
\newblock In \emph{Proceedings of the 54th Annual Meeting of the Association for Computational Linguistics (Volume 1: Long Papers)}, pages 1715--1725, Berlin, Germany. Association for Computational Linguistics.

\bibitem[{Ueda et~al.(2023)Ueda, Ishii, and Miyao}]{ueda2023on}
Ryo Ueda, Taiga Ishii, and Yusuke Miyao. 2023.
\newblock \href {https://openreview.net/forum?id=b4t9_XASt6G} {On the word boundaries of emergent languages based on harris's articulation scheme}.
\newblock In \emph{The Eleventh International Conference on Learning Representations}.

\bibitem[{Ueda and Taniguchi(2024)}]{DBLP:conf/iclr/UedaT24}
Ryo Ueda and Tadahiro Taniguchi. 2024.
\newblock \href {https://openreview.net/forum?id=HC0msxE3sf} {Lewis's signaling game as beta-vae for natural word lengths and segments}.
\newblock In \emph{The Twelfth International Conference on Learning Representations, {ICLR} 2024, Vienna, Austria, May 7-11, 2024}. OpenReview.net.

\bibitem[{Ueda and Washio(2021)}]{ueda-washio-2021-relationship}
Ryo Ueda and Koki Washio. 2021.
\newblock \href {https://doi.org/10.18653/v1/2021.acl-srw.6} {On the relationship between {Z}ipf`s law of abbreviation and interfering noise in emergent languages}.
\newblock In \emph{Proceedings of the 59th Annual Meeting of the Association for Computational Linguistics and the 11th International Joint Conference on Natural Language Processing: Student Research Workshop}, pages 60--70, Online. Association for Computational Linguistics.

\bibitem[{Vithanage et~al.(2023)Vithanage, Wijesinghe, Xavier, Tissera, Jayasena, and Fernando}]{vithanage}
Kasun Vithanage, Rukshan Wijesinghe, Alex Xavier, Dumindu Tissera, Sanath Jayasena, and Subha Fernando. 2023.
\newblock \href {https://doi.org/10.1371/journal.pone.0295748} {Accelerating language emergence by functional pressures}.
\newblock \emph{PLOS ONE}, 18(12):1--28.

\bibitem[{Williams(1992)}]{reinforce}
Ronald~J. Williams. 1992.
\newblock \href {https://doi.org/10.1007/BF00992696} {Simple statistical gradient-following algorithms for connectionist reinforcement learning}.
\newblock \emph{Mach. Learn.}, 8(3–4):229–256.

\bibitem[{Wu et~al.(2019)Wu, Cotterell, and O{'}Donnell}]{wu-etal-2019-morphological}
Shijie Wu, Ryan Cotterell, and Timothy O{'}Donnell. 2019.
\newblock \href {https://doi.org/10.18653/v1/P19-1505} {Morphological irregularity correlates with frequency}.
\newblock In \emph{Proceedings of the 57th Annual Meeting of the Association for Computational Linguistics}, pages 5117--5126, Florence, Italy. Association for Computational Linguistics.

\bibitem[{Zhang(2025)}]{zhang-2025-combinatorial}
Zheyuan Zhang. 2025.
\newblock \href {https://aclanthology.org/2025.coling-main.112/} {A combinatorial approach to neural emergent communication}.
\newblock In \emph{Proceedings of the 31st International Conference on Computational Linguistics}, pages 1660--1666, Abu Dhabi, UAE. Association for Computational Linguistics.

\bibitem[{Zuidema and de~Boer(2009)}]{zuidema}
Willem Zuidema and Bart de~Boer. 2009.
\newblock \href {https://www.sciencedirect.com/science/article/pii/S0095447008000624#aep-bibliography-id30} {The evolution of combinatorial phonology}.
\newblock \emph{Journal of Phonetics}, 37:125--144.

\end{thebibliography}
